\documentclass[12pt,letterpaper]{article}
\usepackage{algorithm}
\usepackage{algpseudocode}
\usepackage{amsfonts,amsmath}

\usepackage[hidelinks]{hyperref}
\usepackage{cleveref}
\crefname{equation}{Eq.}{Eqs.}
\crefname{figure}{Fig.}{Figs.}
\usepackage{mathtools}

\usepackage[mathscr]{eucal}
\usepackage{eucal}
\usepackage{amsmath,amssymb,amsfonts}
\usepackage{microtype}
\usepackage{graphicx}
\usepackage{subfigure}
\usepackage{booktabs} 
\usepackage{float}
\usepackage[numbers]{natbib}

\usepackage{authblk}

\usepackage{amsthm}
\usepackage{placeins}
\usepackage{url}

\newcommand{\sign}{\operatorname{sign}}
\newcommand{\calD}{{\cal D}} 
\newcommand{\calP}{{\cal P}} 
\newcommand{\calU}{{\cal U}} 
\newcommand{\calS}{{\cal S}} 
\newcommand{\reals}{{\mathbb{R}}} 
\newcommand{\E}{{\mathbb{E}}} 
\newcommand{\PU}{{\rm PU}}
\newcommand{\pp}{{\rm p}} 
\newcommand{\nn}{{\rm n}} 
\newcommand{\uu}{{\rm u}} 
\DeclareMathOperator*{\argmin}{\arg\!\min}
\usepackage{bbm}
\def \bbI {\mathbbm{I}}
\usepackage{algorithm}
\usepackage{algpseudocode}

\newcounter{cmt}

\usepackage{caption}
\usepackage{geometry}
\usepackage{multirow}

\theoremstyle{definition}

\usepackage{amsfonts}
\usepackage{bbm}
\def \bbI {\mathbbm{I}}

\usepackage{xspace}

\newcommand{\codename}{AdaPU\xspace}

\AtBeginDocument{%
  \providecommand\BibTeX{{%
    \normalfont B\kern-0.5em{\scshape i\kern-0.25em b}\kern-0.8em\TeX}}}

\begin{document}

\title{A Boosting Algorithm for Positive-Unlabeled Learning}

\author[1,$\dag$]{Yawen Zhao}

\author[1,$\dag$]{Mingzhe Zhang}

\author[1]{Chenhao Zhang}

\author[2]{Weitong Chen}

\author[1]{Nan Ye}
 
\author[1,3*]{Miao Xu}

\affil[1]{The University of Queensland, \authorcr \{yawen.zhao,mingzhe.zhang,chenhao.zhang,nan.ye,miao.xu\} \newline @uq.edu.au}
\affil[2]{The University of Adelaide, \authorcr {t.chen@adelaide.edu.au}}
\affil[3]{RIKEN Japan}
\affil[$\dag$]{These authors contributed equally to this work.}
\date{}
\maketitle

\begin{abstract}

Positive-unlabeled (PU) learning deals with binary classification problems when only positive (P) and unlabeled (U) data are available. 
Many recent PU methods are based on neural networks, but little has been done to
develop boosting algorithms for PU learning, despite boosting algorithms' strong
performance on many fully supervised classification problems. 
In this paper, we propose a novel boosting algorithm, \codename, for PU learning.
Similarly to AdaBoost, \codename aims to optimize an empirical exponential loss,
but the loss is based on the PU data, rather than on positive-negative (PN) data.
As in AdaBoost, we learn a weighted combination of weak classifiers by learning
one weak classifier and its weight at a time.
However, \codename requires a very different algorithm for learning the weak
classifiers and determining their weights.
This is because \codename learns a weak classifier and its weight using a
weighted positive-negative (PN) dataset with some \emph{negative} data weights
---
the dataset is derived from the original PU data, and the data weights are
determined by the current weighted classifier combination, but some data weights
are negative.
Our experiments showed that \codename~outperforms neural networks on several
benchmark PU datasets, including a large-scale challenging cyber security dataset.

\end{abstract}

\section{Introduction}\label{introduction}

Positive-unlabeled (PU) learning~\cite{denis1998pac, comite1999positive, denis2005learning} has recently attracted significant interest as an important machine learning problem.
While traditional supervised binary classification considers learning from both positive and negative examples, 
PU learning considers learning from only positive and unlabeled examples, where an unlabeled example can be either positive (P) or negative (N). 
PU learning arises naturally in various domains where positive (P) data and unlabeled (U) data are readily available, but obtaining negative examples is costly. 
For example, in disease diagnosis, we can easily collect data of confirmed patients provided by the doctors, but not for patients with mild or asymptomatic symptoms that have not been diagnosed yet. 

Various approaches have been proposed for PU learning. 
One common approach is an iterative two-step approach:
in each iteration, the current model is used to identify reliable N data from U, then ordinary fully-supervised (PN) learning is performed to update the current model~\cite{liu2002partially, li2003learning}. 
Another important approach performs supervised learning on a weighted PN dataset derived from the PU dataset \cite{DBLP:conf/kdd/ElkanN08,du2015,kiryo2017positive}.
Our work follows this second approach, and is partly motivated by unbiased PU (uPU)~\cite{du2015} and its improved variant, non-negative PU (nnPU)~\cite{kiryo2017positive}.

Recent state-of-the-art (SOTA) PU methods are mostly based on neural networks
(NNs)~\cite{kiryo2017positive, DBLP:chen2020self, su2021positive},
due to the recent successes of NNs in various domains, particularly in computer vision and natural language processing.
On the other hand, boosting algorithms are still often used as a preferred method for
handling tabular data due to their simplicity and superior performance.
Notably, many datasets on Kaggle are tabular, and boosting algorithms are among one of the top choices of Kaggle users.
\footnote{This is reflected by the $25,973$ responses to the 2021 Kaggle Data Science \& Machine Learning Survey
(\url{https://www.kaggle.com/kaggle-survey-2021}).}
Despite the strong performance of boosting methods in supervised learning,
little has been done to develop boosting algorithms for PU learning yet.
We thus propose to fill this gap.

In this paper, we propose a novel boosting algorithm, \codename, for learning from PU data.
Our algorithm is partly motivated by the classical AdaBoost (Adaptive Boosting)~\cite{Freund1995, adaboostfreund1999short} algorithm.
As in AdaBoost, we learn a number of weak classifiers and their weights
sequentially, and use the weighted combination of them as the final model.
Each weak classifier is trained to do well on examples that are hard for previous weak classifiers;
specifically, the training set of each weak classifier consists of weighted examples, with harder examples having higher weights.
However, while AdaBoost minimizes the empirical exponential loss based on PN data, 
\codename minimizes an empirical exponential loss based on PU data.
This results in some negative weights in the weighted samples and entails a
novel weak learning algorithm:
instead of simply training the weak classifiers to minimize weighted classification errors, 
the training procedure needs to be augmented with a mechanism to prevent overfitting.

The remainder of this paper is organized as follows. 
\Cref{Formulation and Background} briefly reviews AdaBoost and PU learning. 
\Cref{ada-PU section} describes our \codename algorithm. 
The experimental results are presented in \Cref{experiment}, with the conclusion in \Cref{Conclusions}.

\section{Preliminaries}\label{Formulation and Background}

Our \codename algorithm builds on the AdaBoost algorithm and some PU-based risk
estimators.
We provide a brief description of these important ideas below.

\subsection{AdaBoost}\label{related_work_adaboost}
\label{relwork}

Boosting is one of the major types of ensemble methods to ``boost'' a series of weak classifiers outputted by a weak learner to one strong classifier. Roughly speaking, a weak classifier is one that is only guaranteed to be better than random guessing,
while a strong classifier is one that can achieve arbitrarily accurate performance.
AdaBoost (Adaptive Boosting) is one of the most classical boosting algorithms~\cite{Freund1995, adaboostfreund1999short}, with a lot of  successful applications, 
such as face recognition~\cite{viola2004robust}.

We provide a brief review of AdaBoost based on \cite{hastie2009elements}.
Given training data $\{(x_{1}, y_{1}), \ldots, (x_{n}, y_{n})\}$, where $x_i$ is the $i$-th instance and $y_{i} \in \{-1, +1\}$ is the label for $x_i$, AdaBoost aims to find a strong classifier of the form
\begin{align}
    F_{T}(x) \coloneqq \sign\left(H_{T}(x)\right) =\sign\left(\sum_{t=1}^{T} \alpha_{t} h_{t}(x)\right),
\end{align}
where $h_{t}$ is the $t$-th weak classifier, and $\alpha_{t} > 0$ for all $t\in\{1, \ldots, T\}$.

Instead of finding $F_{T}(x)$ that directly minimizes the classification error based on zero-one loss, AdaBoost uses the exponential loss as the surrogate loss and minimizes 
\begin{equation}\label{eq:exp_loss}
	\sum_{i=1}^{n} \exp(- y_{i} H_{T}(x_{i}))
\end{equation}
in a greedy manner by gradually adding newly learned weak classifiers. 
Specifically, AdaBoost starts with $H_{0}(x) = 0$.
At the $t$-th iteration, given $H_{t-1}$, the weak classifier $h_{t}$ and its weight $\alpha_{t}$ are minimizers of 
\begin{equation}
\label{eq:obj_exploss}
       \sum_{i = 1}^{n} e^{-y_i (H_{t-1} (x_i) + \alpha h (x_i))}.
\end{equation}
Let $w^{(t)}_{i} = e^{-y_{i} H_{t-1}(x_{i})}/ \sum_{j}  e^{-y_{j} H_{t-1}(x_{j})}$, and 
$\epsilon_{t}(h) = \sum_{i} w^{(t)}_{i} \bbI(y_{i} \neq h(x_{i}))$ be a weak classifier $h$'s weighted classification error, then 
$h_{t}$ and $\alpha_{t}$ are given by \cite{hastie2009elements}:
\begin{align} \label{eq:slt_hx}
	h_{t} &= \argmin_{h} \epsilon_{t}(h), \\
	\alpha_{t} &= \frac{1}{2} \log \frac{1-\epsilon_{t}}{\epsilon_{t}}.
\end{align}
At each iteration, we can efficiently calculate the weights $w^{(t)}_{i}$ using the recursive formula 
\begin{equation}
    \label{eq:slt_w_update}
		w^{(t)}_{i} = w^{(t-1)}_{i} e^{-\alpha_t y_i h_{t-1}(x_{i})} / Z_{t-1},
\end{equation}
where $Z_{t-1} = \sum_{i} w^{(t-1)}_{i} e^{-\alpha_t y_i h_{t-1}(x_{i})}$ is the normalization constant.

\subsection{PU learning}\label{pu learning background}

The study of PU learning can be traced back to~\cite{denis1998pac, comite1999positive, denis2005learning}. 
Early works are usually based on sample selection, which first uses heuristics to select reliable N data from U and then performs PN learning~\cite{liu2002partially, li2003learning, liu2003building}. 
Various learning algorithms, including boosting, can be used in the supervised learning step.
The two-step method's performance can be sensitive to the selection strategy, and a poor selection strategy may lead to unsatisfactory performance.
To the best of our knowledge, the only boosting method for PU data is such a two-step method~\cite{zhang2018boosting}.
Instead of using a PN boosting algorithm as a building block, our \codename algorithm is designed as a boosting algorithm that directly learns from PU data.

Besides these two-step methods, another group of methods are based on minimizing loss on a weighted PN dataset
constructed using the PU data.
The construction of the weighted PN dataset depends on how PU data is generated.
Two data generation mechanisms are often considered in the literature~\cite{DBLP:conf/icml/MenonROW15}: 
censored PU learning (e.g., \cite{DBLP:conf/kdd/ElkanN08}) where data are sampled together and part of P and all of N data become unlabeled, 
and case-control PU learning (e.g., \cite{lee2003learning,du2014analysis}), where P and U data are sampled separately. 
This paper considers case-control PU learning.
Two recent methods in this category are most closely related to our work: unbiased PU learning (uPU)~\cite{du2015} and non-negative PU learning (nnPU)~\cite{kiryo2017positive}.
uPU allows computing unbiased risk estimates using PU data only.
However, uPU suffers from severe overfitting when utilizing neural networks (NN).
Non-negative PU learning (nnPU)~\cite{kiryo2017positive} was proposed as an improved variant to alleviate overfitting, and it has been adopted in multiple
NN-based methods~\cite{DBLP:chen2020self, su2021positive, chen2021cost} since
then.

We describe uPU and nnPU in detail below.
Let $X \in \mathbb{R}^d$ and $Y \in {\{+1, -1\}}$ be the input and output random
variables respectively,
$p(x) = P(X)$ be the marginal distribution of the input,
$p_{+}(x) = P(x\vert Y = +1)$ be the distribution of positive examples, 
$p_{-}(x) = P(x \vert Y = -1)$ be the distribution of negative examples,
$\pi = \pi_{\rm p} = P(Y = +1)$ be the probability of positive examples, 
$\pi_{\rm n} = 1 - \pi_{\rm p}$ be the probability of negative examples, 
and $\ell: \mathbb{R} \times \{+1, -1\} \rightarrow \mathbb{R}$ be a loss function
with $\ell(y', y)$ being the loss incurred when an example in class $y$ is
predicted to have a score $y'$.
Following~\cite{du2014analysis, kiryo2017positive, DBLP:hsieh2019classification,
DBLP:chen2020self}, we assume $\pi_{\rm p}$ is known throughout the paper. 
As in the fully supervised case, the objective of PU learning is to find a
classifier $g: \mathbb{R}^{d} \to \mathbb{R}$ minimizing the expected risk
\begin{align}
	R(g) = \mathbb{E}(\ell(g(X), Y)).
\end{align}

In fully-supervised classification, we have a sample of training data drawn
independently from $P(X, Y)$, and this allows us to use the average loss to
estimate the expected risk.
However, in PU learning, we only have access to positive and unlabeled examples,
but not negative examples, thus estimating the expected risk using the available
data is non-trivial. 
To address such a difficulty, the risk is rewritten as the following using PU data~\cite{du2014analysis,du2015}
\begin{align*}
	R(g)
	&=
	\pi_{\rm p} \mathbb{E}_{p_{+}(x)}[\ell(g(X), +1)] 
	+
	\left(\mathbb{E}_{p(x)}[\ell(g(X), -1)] - \pi_{\rm p}\mathbb{E}_{p_{+}(x)}[\ell(g(x), -1)]\right). 
\end{align*}
If our PU data consists of a set $\calP = \{x^{+}_{1}, \ldots, x^{+}_{n_{\rm p}}\}$ of positive
examples sampled independently from $p_{+}(x)$, and a set 
$\calU = \{x^{{\uu}}_{1}, \ldots, x^{{\uu}}_{n_{\uu}}\}$ of unlabeled
examples sampled independently from $p(x)$, we have the following unbiased risk estimator \cite{du2014analysis,du2015}
\begin{eqnarray}\label{eqn:pu-empirical-general}
	\widehat{R}_{\rm pu}(g) 
	= 
	\frac{\pi}{n_{\rm p}}\sum_{x\in {\calP}}\ell(g(x), +1) + 
	\frac{1}{n_{\uu}} \sum_{x\in {\calU}} \ell(g(x),\!-\!1) - \frac{\pi}{n_{\rm p}}\sum_{x\in {\calP}}\ell(g(x),\!-\!1).
\end{eqnarray}
uPU~\cite{du2015} optimizes the above unbiased risk estimator.
It is observed to easily overfit when using highly expressive models such as deep neural networks~\cite{kiryo2017positive}.
This is because the difference of the second and third term in the unbiased risk estimator is an estimate of the non-negative loss
$\mathbb{E}_{p_{-}}[\ell(g(x), -1)]$, but it can be negative.
To alleviate overfitting, the nnPU estimator thresholds the estimate to ensure that it is non-negative~\cite{kiryo2017positive}
\begin{eqnarray*}
		\widehat{R}_{\rm nnpu}(g) 
		&=& \frac{\pi}{n_{\rm p}}\sum_{x\in {\calP}}\ell(g(x), +1) + \\
		&&\max \left( \! 0, \! \frac{1}{n_{\rm u}} \! \sum_{x\in {\calU}} \ell(g(x), -1) \! - \! \frac{\pi}{n_{\rm p}}\sum_{x\in {\calP}}\ell(g(x), -1)) \! \right).
\end{eqnarray*}

\section{The \codename Algorithm}\label{ada-PU section}

The overall algorithm of \codename is shown in \Cref{alg:adapu}.
Similarly to AdaBoost, \codename is designed to minimize an empirical
exponential loss. \codename has an iterative structure similar to AdaBoost too:
at each iteration, it learns a weak classifier $h_{t}$ and its weight
$\alpha_{\PU}^{t}$ on a dataset with weights $\{w^{(t)}\}$ using \Cref{alg:stump}.
We then compute a new aggregate classifier 
$H_{t} = H_{t-1} + \beta \alpha_{\PU}^{t} h_{t}$, where $\beta \in (0, 1)$ acts
as a regularization constant.
The weights $w^{(t)}$ are then updated to new weights $w^{(t+1)}$. 

\begin{algorithm}[t!]
\caption{\codename}\label{alg:adapu}
\hspace*{\algorithmicindent} \textbf{Input} $\pi$; $\calP$; $\calU$; $T$; $\beta$; $K$ \\
\hspace*{\algorithmicindent} \textbf{Output} $F_T (x)$
\begin{algorithmic}[1]
	\Procedure{\codename}{}
	\State $H_0 \gets 0$, and initialize $w^{(1)}$ according to \Cref{eq:w0}.
	\For {$t=1,\ldots, T$}
	\State $h_{t}, \alpha_{t} \gets \text{StumpGenerator}(\pi, w^{(t)}, {\calP}, {\calU}, K)$.
	\State $H_{t} \gets H_{t-1} + \beta\alpha_{t} h_t$.
	\State $w^{(t+1)}(x, y) \gets w^{(t)}(x, y) e^{- \beta \alpha_{t} y h_{t}(x)}$ for each $(x, y)$.
	\EndFor
\State $F_T (x) \gets \operatorname{sign}\left(H_T (x) \right)$.
\EndProcedure
\end{algorithmic}
\end{algorithm}

While \codename and AdaBoost share some similarities at a high level, there is a key difference in their empirical exponential losses that results in important
algorithmic differences.
Specifically, AdaBoost's loss relies on a fully labeled dataset, while
\codename's loss is based on a weighted PN dataset derived from the PU dataset,
where the example weights can be negative.
This results in major differences in how a weak classifier and its weight are
learned in \codename and AdaBoost.

Below, we describe the objective function of \codename, provide a derivation of 
a greedy minimization algorithm for learning a weak classifier and its weight,
and then explain the complete pseudo-code for a concrete instantiation of the
greedy procedure \Cref{alg:stump}.

\paragraph{\codename's empirical exponential loss}
For any classifier $H: \reals^{d} \to \reals$, \codename estimates its expected
exponential loss $\E (e^{-Y H(X)})$ using the uPU estimator:
\begin{align}
	L_{\exp}(H)
	=
	\frac{\pi}{n_{\pp}}
	\sum_{x \in {\calP}} e^{-H(x)}
	+
	\frac{1}{n_{\uu}}
	\sum_{x \in {\calU}} e^{H(x)}
	- 
	\frac{\pi}{n_{\pp}}
	\sum_{x \in {\calP}} e^{H(x)}
\end{align}
This can be written down as the exponential loss of $H$ on a weighted PN dataset
$\calD$ derived from the PU data:
\begin{align}
	L_{\exp}(H)
	=
	\sum_{(x, y) \in \calD} w^{(1)}(x,y) e^{-y H(x)},
\end{align}
where $\calD = \calP^{+} \cup \calP^{-} \cup \calU^{-}$
is the union of three fully labeled datasets 
${\calP}^+ = \{(x,+1): x\in \calP\}$, 
${\calP}^- = \{(x,-1): x\in \calP\}$, and
${\calU}^- = \{(x,-1): x\in \calU\}$.
The weights $w^{(1)}$ are defined by
\begin{equation}\label{eq:w0}
    w^{(1)}(x,y) 
		= 
		\begin{cases}
			\frac{\pi}{n_{\pp}}, & (x, y) \in {\calP}^{+} \\
			\frac{1}{n_{\uu}}, &  (x, y) \in {\calU}^{-}\\
			-\frac{\pi}{n_{\pp}}, & (x, y) \in {\calP}^{-}.
		\end{cases}
\end{equation}
Note that for $(x, y) \in \calP^{-}$, the weight $w^{(1)}(x, y)$ is negative.

\paragraph{Derivation of a greedy minimization algorithm}

\codename uses an iterative greedy algorithm to construct a classifier 
$H_{T}(x) = \sum_{t=1}^{T} \alpha_{t} h_{t}(x)$, where each $h_{t}$ is a weak
classifier and each $\alpha_{t} > 0$.
It starts with an initial classifier $H_{0} = 0$.
At iteration $t$, given $H_{t-1}$, \codename aims to choose $h_{t}$ and
$\alpha_{t}$ to further reduce the exponential loss
\begin{align}
	L_{t}(h, \alpha) 
	=
	\sum_{(x, y) \in \calD} w^{(1)}(x,y) e^{-y [H_{t-1}(x) + \alpha h(x)]}.
\end{align}
Let $w^{(t)}(x, y) = w^{(1)}(x, y) e^{-y H_{t-1}(x)}$, and 
$Z_{t} = \sum_{(x, y) \in \calD} w^{(t)}(x, y)$ be the total weight.
By grouping examples based on whether they are correctly classified by $h$, we
can rewrite $L_{t}(h, \alpha)$ as 
\begin{align*}
	L_{t}(h, \alpha) 
	&= 
	(Z_{t} - E_{t}(h)) e^{-\alpha}
	+ E_{t}(h) e^{\alpha},
\end{align*}
where $E_{t}(h)$ is the total weight of examples misclassified by $h$:
\begin{align}
	E_{t}(h)
	=
	\sum_{(x, y) \in \calD, h(x) \neq y} w^{(t)}(x,y).
\end{align}

Naively, we can choose $h_{t}$ and $\alpha_{t}$ as the minimizers of
$L_{t}(h, \alpha)$, but this can easily lead to overfitting.
To see this, first note that if $E_{t}(h) > 0$ for any $h$, then the minimizer
of $L_{t}(h, \alpha)$ can be found by taking $h$ as the minimizer of $E_{t}$,
and then setting
$\alpha = \ln \frac{Z_{t}-E_{t}(h)}{E_{t}(h)}$.
However, $E_{t}(h)$ may be negative, because the weight
$w^{(t)}(x,y)$ for $(x, y) \in \calP^{-}$ is negative --- in fact, the weight
becomes a large negative number if previous weak classifiers correctly classify
the positive example $x$ and classify the unlabeled examples as negative.
When $E_{t}(h)$ is negative, the loss takes a minimum value of $-\infty$ when $\alpha =
\infty$, and the ensemble is thus dominated by $h$ alone. 
This easily leads to overfitting as even the best weak classifier may have low accuracy.

We introduced a regularization mechanism motivated by nnPU by requiring the estimated error
of the weak classifier $h$ on negative examples to be non-negative.
Specifically, consider the weighted classification error $e_{t}(h) = E_{t}(h)/Z_{t}$ and
$e_{t}^{nn}(h) = \sum_{(x, y) \in \calP^{-}\cup \calU^{-}, h(x) \neq y} w^{(t)}(x,y)/Z_{t}$.
We interpret $e_{t}^{nn}(h)$ as an estimate for the expected error of $h$ on the negative examples.
Then we choose $h_{t}$ by minimizing $E_{t}(h)$ under the constraints that 
$e_{t}(h) \in [0, 0.5)$ and $e_{t}^{nn}(h) \ge 0$.
The $\alpha$ value is chosen to minimize $L(h_{t}, \alpha)$, giving us 
$\alpha_{t} = \frac{1}{2} \ln \frac{1 - e_{t}(h_{t})}{e_{t}(h_{t})}$.
In summary, we have
\begin{align}
	h_{t} &= \argmin_{h: e_{t}(h) \in [0, 0.5), e_{t}^{nn}(h) \ge 0} E_{t}(h), \label{eq:ht_overall}\\
	\alpha_{t} &= \frac{1}{2} \ln \frac{1 - e_{t}(h_{t})}{e_{t}(h_{t})}.  \label{eq:at_overall}
\end{align}
The condition $e_{t}(h) \in [0, 0.5)$ can be interpreted as requiring $h$ to be 
better than random guessing, and it ensures that the weight $\alpha_{t}$ is positive.
The condition $e_{t}^{nn}(h) > 0$ rules out classifiers with a negative estimate for the 
expected error on the negative examples.
Note that $Z_{t}$ may be less than or equal to 0.
We stop adding weak classifiers whenever this happens.

We note that in PU learning, we often have a lot of unlabeled examples, and thus a
large part of the weighted error $e_{t}(h)$ is contributed by the unlabeled examples.
The error $e_{t}(h)$ can be relatively small when most of the unlabeled examples are 
classified as negative, but a small fraction of the positive examples are classified as
positive. 
However, since the positive examples are indeed positive, while the unlabeled examples 
may be either positive or negative, it is more important to correctly classify the 
positive examples as positive.
We thus introduce an error measure $\epsilon_{t}(h)$ that better balance the influence 
of the positive examples and the unlabeled examples, as compared to $e_{t}(h)$.
Specifically, for any subset $\calS$ of $\calD$, let 
$w^{(t)}(\calS) = \sum_{(x, y) \in \calS} w^{(t)}(x, y)$, and 
$\epsilon_{t}(h; \calS)
=
\frac{1}{w^{(t)}(\calS)} \sum_{(x, y) \in \calS} w^{(t)}(x, y) \bbI(h(x) \neq y)$.
We define $\epsilon_{t}(h)$ by
\begin{align*}
	\epsilon_{t}(h)
	&=
	\pi \epsilon_{t}(h; \calP^{+})
	+
	\epsilon_{t}(h; \calU^{-})
	- 
	\pi \epsilon_{t}(h; \calP^{-}).
\end{align*}
The form is inspired by the uPU estimator for expected zero-one loss, which is
\begin{align*}
	L_{01}(h)
	=
	\pi
	\sum_{x \in {\calP}} \frac{\bbI(h(x) \neq +1)}{n_{\pp}}
	+
	\sum_{x \in {\calU}} \frac{\bbI(h(x) \neq -1)}{n_{\uu}}
	- 
	\pi
	\sum_{x \in {\calP}} \frac{\bbI(h(x) \neq -1)}{n_{\pp}} 
\end{align*}

Finally, we choose the weak classifier $h_{t}$ and its weight $\alpha_{t}$ by
\begin{align}
	h_{t} &= \argmin_{h: \epsilon_{t}(h) \in [0, 0.5), \epsilon_{t}^{nn}(h) \ge 0} E_{t}(h), \label{eq:ht}\\
	\alpha_{t} &= \frac{1}{2} \ln \frac{1 - \epsilon_{t}(h_{t})}{\epsilon_{t}(h_{t})}.  \label{eq:at}
\end{align}
where $\epsilon_{t}^{nn} = \epsilon_{t}(h; \calU^{-}) - \pi \epsilon_{t}(h; \calP^{-})$
is the counterpart of $e_{t}^{nn}$.

We also implement an additional regularization mechanism by adding $\beta \alpha_{t} h_{t}$ 
to $H_{t-1}$ for some $\beta \in [0, 1]$, instead of adding $\alpha_{t} h_{t}$ to 
$H_{t-1}$.
This is motivated by the regularization mechanism in \cite{mannor2003greedy},
which is implemented in scikit-learn.

We provide a comparison between the updates in \Cref{eq:ht_overall,eq:at_overall} (which 
are said to use \emph{over-all normalization}) and the updates in \Cref{eq:ht,eq:at} (which are
said to use \emph{per-group normalization}), in \Cref{ablation study}.
Both variants have similar performance with a large number of weak classifiers, but per-group
normalization seems to learn faster at the beginning. 
This suggests that balancing the contribution of the errors from $\calP^{+}, \calP^{-}$ and $\calU^{-}$
is helpful.

\paragraph{An instantiation of the greedy minimization algorithm}

We show how the greedy minimization approach discussed above can be instantiated
when the weak classifiers are decision stumps.
Decision stumps allow an efficient algorithm for the optimization problem in
\Cref{eq:ht}, yet shows strong empirical performance in our experiments.
Our algorithm is shown in \Cref{alg:stump}.
For computational efficiency, we do not consider all possible thresholds for all
the features for the decision stumps considered.
Instead, for each feature, we consider $K$ random thresholds sampled in an interval that is slightly larger than the feature's value range --- note that a larger range is chosen to allow decision stumps that classify all examples as
positive/negative.
We then find $h_{t}$ by considering only these candidate decision stumps.
If all the candidate decision stumps do not satisfy the conditions 
$\epsilon_{t}(h) \in [0, 0.5)$ and $\epsilon_{t}^{nn}(h) \ge 0$, then
$h_{t} = 0$ with $\alpha_{t} = 0$, thus $H_{t}$ is the same as $H_{t-1}$.
Interestingly, this did not happen in our experiments.

\begin{algorithm}[t!]
	\caption{\codename~StumpGenerator}\label{alg:stump}
	\hspace*{\algorithmicindent} \textbf{Input} $\pi$; $w^{(t)}$; $\calP$; $\calU$; $K$ \\
	\hspace*{\algorithmicindent} \textbf{Output} $h_{t}$, $\alpha_{t}$
	\begin{algorithmic}[1]
		\Procedure{StumpGenerator}{}
		\State $E_{\min} \gets \infty$; $\epsilon_{min} \gets 1/2$; $h_{t} \gets 0$ 
		\For {each feature $f$}
			\State $n_{f} \gets$ the number of $f$'s unique values.
			\State $v_{r} \gets v_{\max} - v_{\min}$, where $[v_{\min}, v_{\max}]$ is $f$'s range.
			\For{$k = 1,\ldots, K$} 
				\State $v \gets $ a random value in 
					$\left[v_{\min} - \frac{v_r}{(n_f - 1)}, v_{\min} + \frac{v_r}{n_f - 1}\right]$.
				\State $h_{L}$, $h_{R}$ $\gets$ the two decision stumps splitting $f$ with threshold $v$.
				\For{$h$ in $[h_{L}, h_{R}]$}
					\If {$\epsilon_{t}(h) \in [0, 0.5)$ and $\epsilon_{t}^{nn}(h) \ge 0$ and $E_{t}(h) < {E}_{\min}$}
							\State $ {E}_{\min} \gets  {E}_{t}$; $h_t \gets h$; $\epsilon_{\min} \gets  \epsilon_t$
					\EndIf
				\EndFor
			\EndFor
		\EndFor
		\State $\alpha_{t} = 1/2\ln [(1-\epsilon_{\min})/\epsilon_{\min}]$ 
		\EndProcedure
	\end{algorithmic}
\end{algorithm}     

\section{Experiments}\label{experiment}
We perform experiments to compare \codename with 
(i) NN-based PU methods, and 
(ii) fully supervised boosting models. 
We also perform experiments to study how the performance of \codename is
affected by the value of the regularization constant $\beta$, different
feature threshold sampling methods and different normalization methods.

\subsection{Datasets} \label{sec:Dataset}
We used PU datasets and PN datasets derived from the following four
datasets.

 \begin{itemize}
    \item
			Epsilon\footnote{\url{https://www.csie.ntu.edu.tw/~cjlin/libsvmtools/datasets/binary.html}}:
			a binary classification text dataset with $400,000$ training and $100, 000$ test examples, each with $2,000$ features. 
			We use $n = 40,000$ training examples from the original dataset. 
			$\pi_{\rm p} = 0.5$. 

    \item Breast Cancer\footnote{\url{https://goo.gl/U2Uwz2}}:
			a binary classification dataset with $n = 455$ training and $114$ test examples. Each example has $30$ features. 
			$\pi_{\rm p}=0.59$.
    \item UNSW-NB15\footnote{\url{https://research.unsw.edu.au/projects/unsw-nb15-dataset}}:
    a binary classification dataset with $n = 175,340$ training and $82,331$ test examples. Each example has $39$ features. $\pi_{\rm p}=0.68$.
    
    \item CIFAR-10\footnote{\url{https://www.cs.toronto.edu/~kriz/cifar.html}}:
			a multi-class dataset with $n = 50,000$ training and $10,000$ test images.
			The input for NN-based methods (i.e., uPU and nnPU methods) are the
			images, and the input for non NN-based methods (i.e., \codename and PN
			boosting methods) are the $3,072$ features provided by a pretrained all
			convolutional net~\cite{DBLP:journals/corr/SpringenbergDBR14}.
			To make it binary, we follow~\cite{kiryo2017positive} treating  classes `airplane', `automobile', `ship' and `truck' as the positive class. 
			$\pi_{\rm p} = 0.4$.
\end{itemize}
 
We follow~\cite{kiryo2017positive} to generate the PU and PN data for our
experiments below.
To generate the PU data, we randomly sample $n_{\pp} = 1000$ positive examples,
and use all training examples as unlabeled examples (i.e., we have $n_{\uu} = n$
unlabeled examples).
To generate the PN data, we randomly sample $n_{\pp} = 1000$ positive examples and 
$n_{\nn}  = (\pi_{\nn} / 2 \pi_{\pp})^2 n_{\pp}$ negative examples.
A strong PU learning algorithm trained on the PU data with a large $n_{\uu}$ is
expected to perform competitively with a PN algorithm trained on such PN
data~\cite{niu2016theoretical}. 
The $n_{\nn}$ values are 256, 55, and 562 respectively for Epsilon, UNSW-NB15
and CIFAR-10; we used $n_{\rm p} = 10$ and $n_{\rm n} = 1$ for the BreastCancer
dataset as it only has $455$ training examples.

\subsection{Experimental settings}
We run each algorithm five times with different random seeds, and report the
means and standard deviations (std) of the performance metrics used.

\medskip
\noindent\textbf{\codename} 
We sampled $K=10$ candidate feature thresholds for each feature.
The regularization constant $\beta$ is an important parameter that needs to be
tuned, as demonstrated in \Cref{ablation study}. 
We used 5-fold cross-validation to choose the best $\beta$ value from 
The selected $\beta$ values are as follows:
$\beta=0.2$ for Epsilon, $\beta=0.1$ for UNSW-NB15, $\beta = 0.001$ for
BreastCancer, and $\beta = 0.2$ for CIFAR-10.

\medskip
\noindent\textbf{PU methods} We compared \codename with neural network based methods uPU~\cite{du2015} and nnPU~\cite{kiryo2017positive}. For Epsilon, BreastCancer and UNSW-NB15, we follow~\cite{kiryo2017positive} to
use \textit{multilayer perceptron}~(MLP), where Epsilon uses
Softsign~\cite{glorot2010understanding} activations and the other two datasets use ReLU~\cite{nair2010rectified} activations. 
For CIFAR-10, we used \textit{Residual Network}~(ResNet)~\cite{he2016deep} and \textit{all convolutional
		net}~\cite{DBLP:journals/corr/SpringenbergDBR14}~(CNN) with
ReLU~\cite{nair2010rectified} activations.
All neural networks are trained using Adam~\cite{kingma2015adam} for $100$
epochs to ensure convergence.

Note that we tune the hyperparameters, including the number of layers for MLP and ResNet, based on the test performance of nnPU and uPU. Specifically, for MLP, the number of hidden layers is chosen from $\{1, 3, 6, 7, 9\}$.
For ResNet, we consider variants with 18, 32, 56, and 110 layers. For CNN, we used the same architecture as in \cite{kiryo2017positive}. For each MLP and ResNet architecture, we tune the hyperparameters learning rate from $\{10^{-2}, 10^{-3}, 10^{-4}\}$ and weight decay from $\{5\times 10^{-8}, 5\times 10^{-9}\}$ and then select the best-performed hyperparameters under that architecture. 
The architectures of nnPU with the best test accuracies are selected: 
ResNet~18 for CIFAR-10, 3-hidden-layer MLP for both Epsilon and BreastCancer, and
9-hidden-layer MLP for UNSW-NB15. While the architectures of uPU with the best test accuracies are: ResNet~110 for CIFAR-10, 3-hidden-layer MLP for BreastCancer, and
9-hidden-layer MLP for both UNSW-NB15 and Epsilon. The selected model architectures and its corresponding hyperparameters are shown in Table~\ref{param_table}.
Note that the test accuracies of these architectures are the same or better than architectures 
chosen using any model selection method, thus the results reported below for nnPU and uPU
are optimistic estimates on their generalization performances.

\medskip
\noindent \textbf{PN methods} We compare our proposed \codename~with XGBoost~\cite{Chen:2016:XST:2939672.2939785}, AdaBoost~\cite{sammearticle} and GBDT~\cite{10.2307/2699986} implemented in scikit-learn using default parameters. The number of iterations is set as $100$, 
which is the same as the number of epochs in neural networks. 

\begin{table}[t]
    \caption{\centering The best model architectures, weight decay, and learning rates for each dataset after hyperparameter tuning and architecture selection for both nnPU and uPU.}
    \label{param_table}
    \footnotesize
    \centering\setlength\tabcolsep{1mm}
    \begin{sc}
        \begin{tabular}{c | c | c | c | c}
        \toprule
         & Dataset & Model Architecture &  Weight Decay  & Learning Rate  \\
        \midrule
        \multirow{4}{1cm}{nnPU} & CIFAR-10 & ResNet 18 &   $5 \times 10^{-8}$  & $1 \times 10^{-4}$  \\
 
        &Epsilon & 3-hidden-layer MLP &  $5 \times 10^{-8}$  & $1 \times 10^{-3}$  \\
        
        &UNSW-NB15 & 9-hidden-layer MLP &  $5 \times 10^{-8}$  & $1 \times 10^{-2}$  \\
        
        &BreastCancer & 3-hidden-layer MLP &  $5 \times 10^{-8}$  & $1 \times 10^{-3}$  \\
        \midrule
        \multirow{4}{1cm}{uPU} & CIFAR-10 & ResNet 110 &   $5 \times 10^{-8}$  & $1 \times 10^{-3}$  \\
 
        &Epsilon & 9-hidden-layer MLP &    $5 \times 10^{-9}$  & $1 \times 10^{-2}$  \\
        
        &UNSW-NB15 & 9-hidden-layer MLP &   $5 \times 10^{-8}$  & $1 \times 10^{-2}$  \\
        
        &BreastCancer & 3-hidden-layer MLP &   $5 \times 10^{-8}$  & $1 \times 10^{-4}$  \\
        \bottomrule
        \end{tabular}
    \end{sc}
\end{table}

\subsection{Results}

\begin{figure}[ht]
    \centering
    \subfigure[CIFAR-10]{
        \label{cnncifar}
        \begin{minipage}[htbp]{0.4\linewidth}
            \centering
            \includegraphics[width=.9\textwidth,height=.6\textwidth]{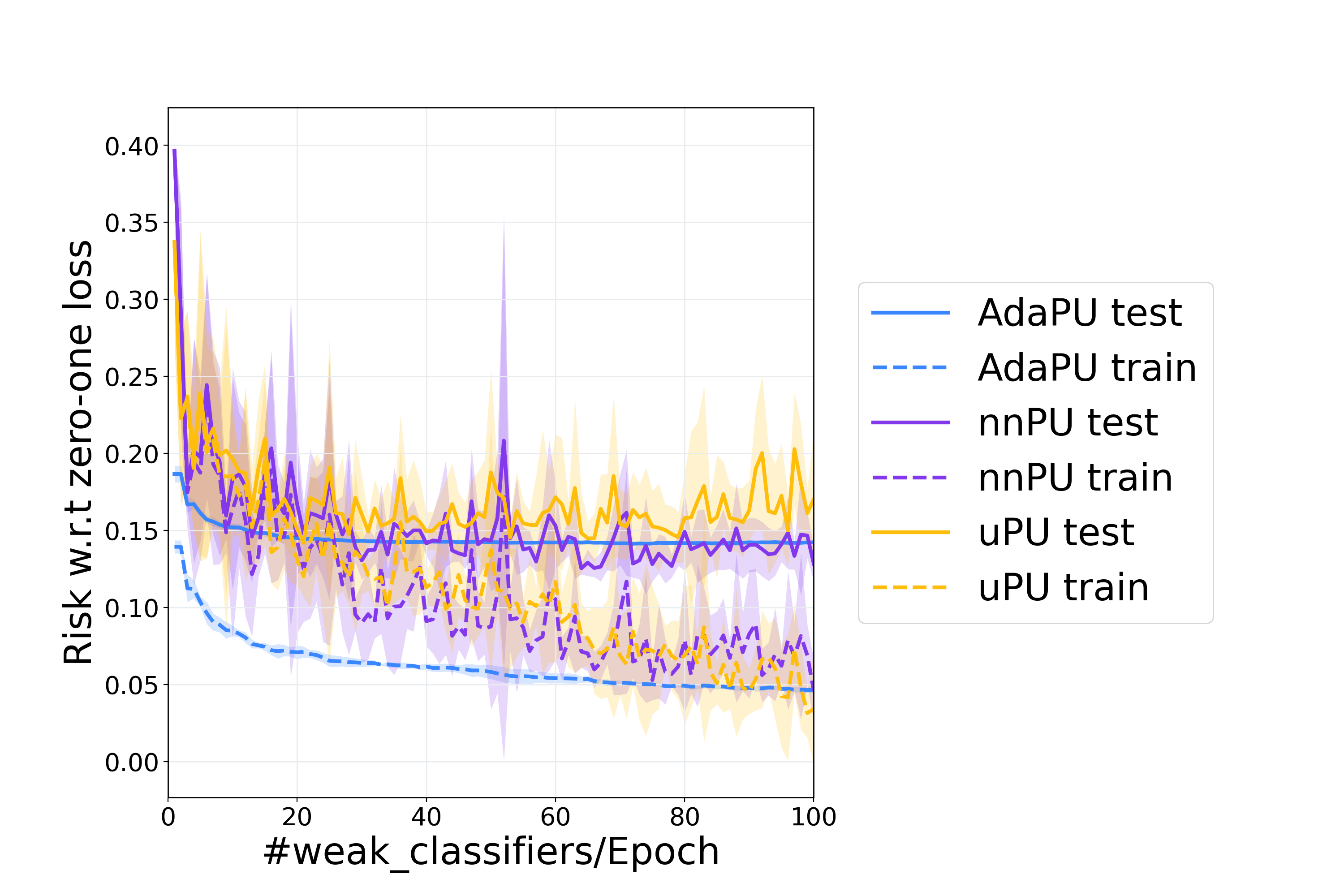}
        \end{minipage}%
    }%
    \subfigure[Epsilon]{
        \begin{minipage}[htbp]{0.4\linewidth}
            \centering
            \includegraphics[width=.9\textwidth,height=.6\textwidth]{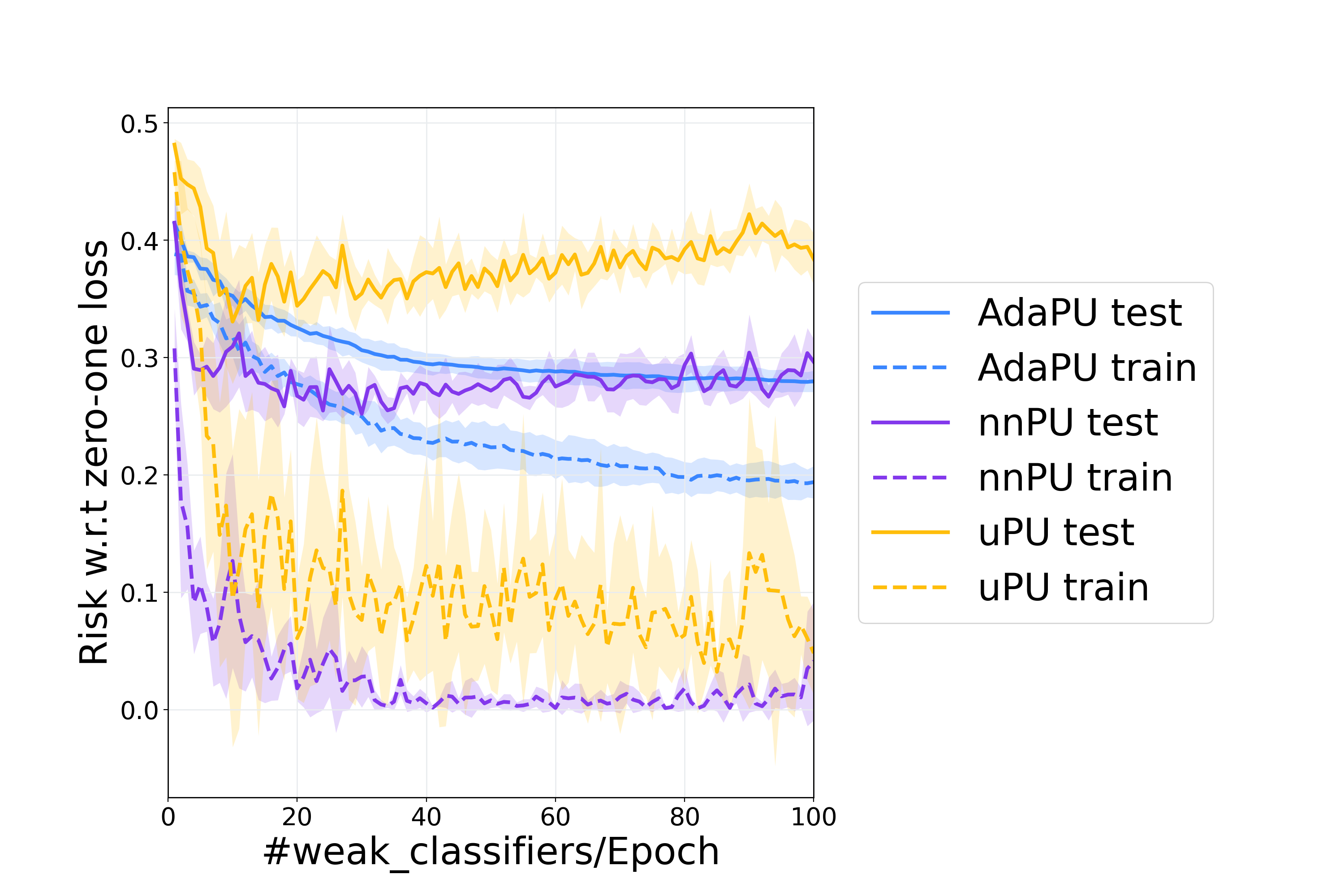}
        \end{minipage}%
    }%
    
    \subfigure[UNSW-NB15]{
        \begin{minipage}[htbp]{0.4\linewidth}
            \centering
            \includegraphics[width=.9\textwidth,height=.6\textwidth]{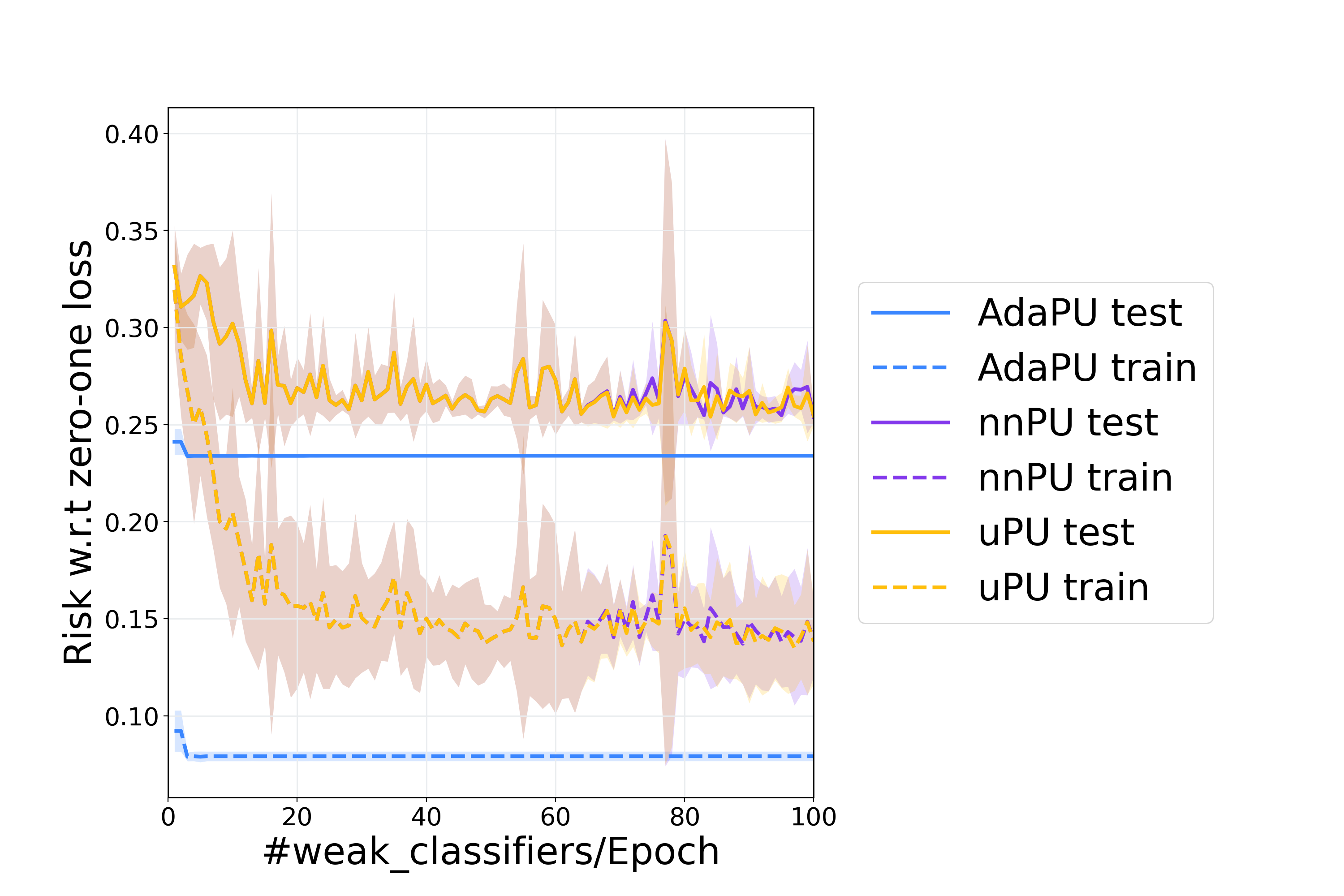}
        \end{minipage}%
    }%
    \subfigure[BreastCancer]{
        \begin{minipage}[htbp]{0.4\linewidth}
            \centering
            \includegraphics[width=.9\textwidth,height=.6\textwidth]{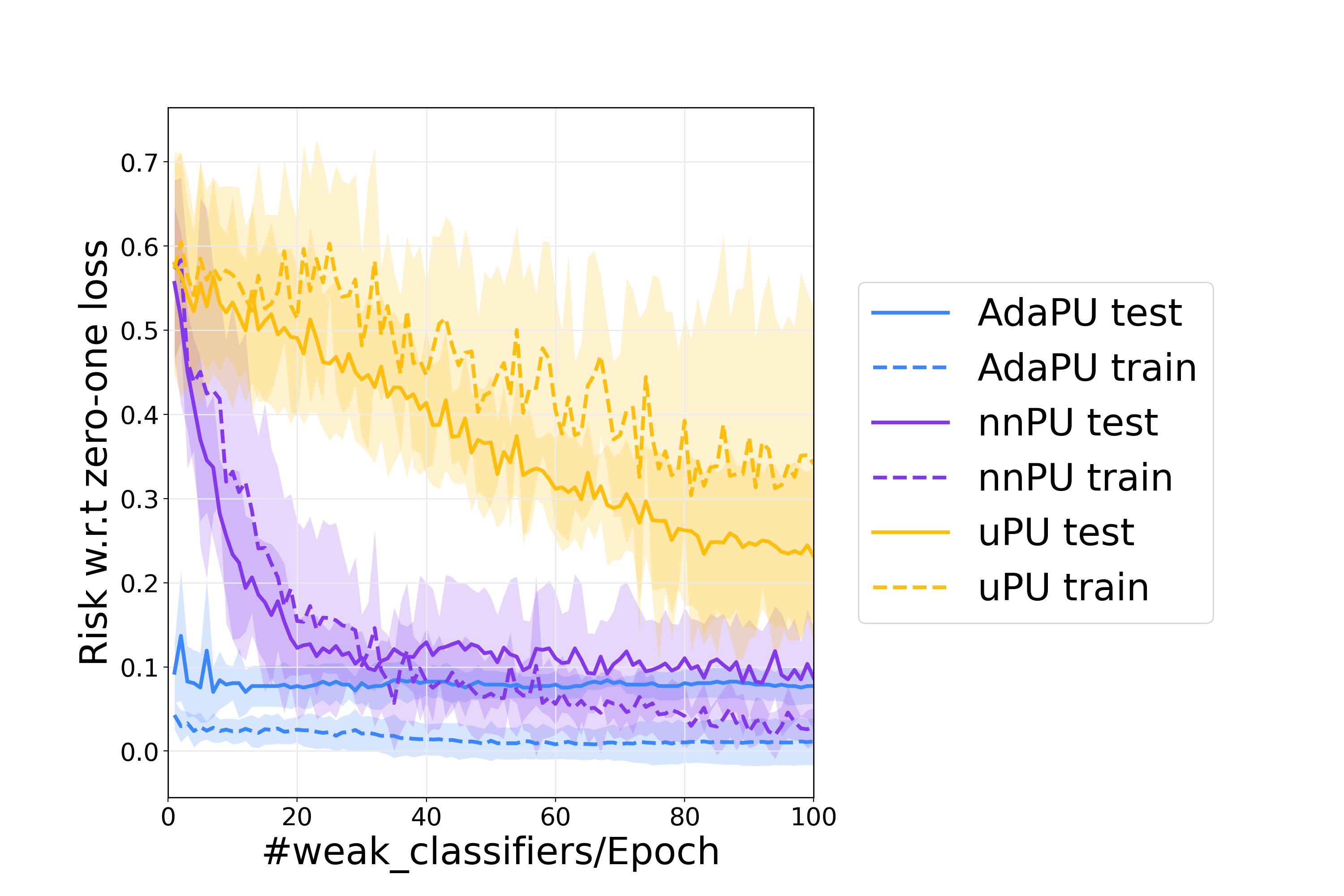}
        \end{minipage}%
    }%
    \centering
    \caption{Training zero-one loss and test zero-one loss of \codename~and NN-based methods, on CIFAR-10, Epsilon, UNSW-NB15, and BreastCancer. 
		The lines and shaded areas show the averages and standard deviations for 5 trials.
		}
    \label{fig:NN-performence}
\end{figure}

\noindent \textbf{Comparison with NN-based PU methods} 
The results of \codename and NN-based PU learning methods are shown in
\Cref{fig:NN-performence} and \Cref{acc_table}. 
\codename outperforms nnPU, and uPU on Epsilon, UNSW-NB15, and BreastCancer. 
\codename performs similarly as nnPU on CIFAR-10. 
In particular, \codename clearly outperformed all NN-based PU learning methods
on tabular data. 

\begin{figure}[ht]
    \centering
    \subfigure[CIFAR-10]{
        \begin{minipage}[htbp]{0.4\linewidth}
            \centering
            \includegraphics[width=.9\textwidth,height=.6\textwidth]{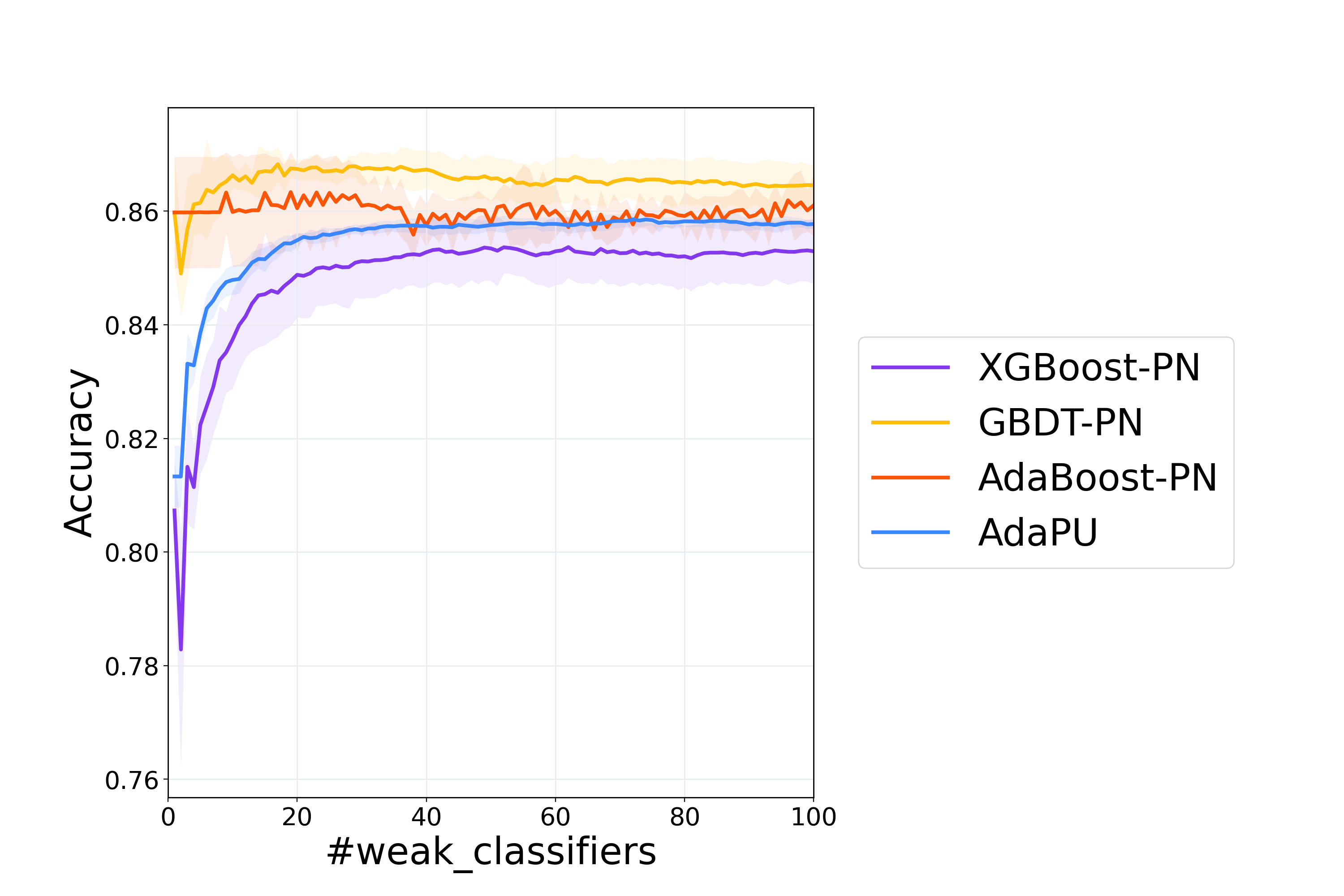}
        \end{minipage}%
    }%
    \subfigure[Epsilon]{
        \begin{minipage}[htbp]{0.4\linewidth}
            \centering
            \includegraphics[width=.9\textwidth,height=.6\textwidth]{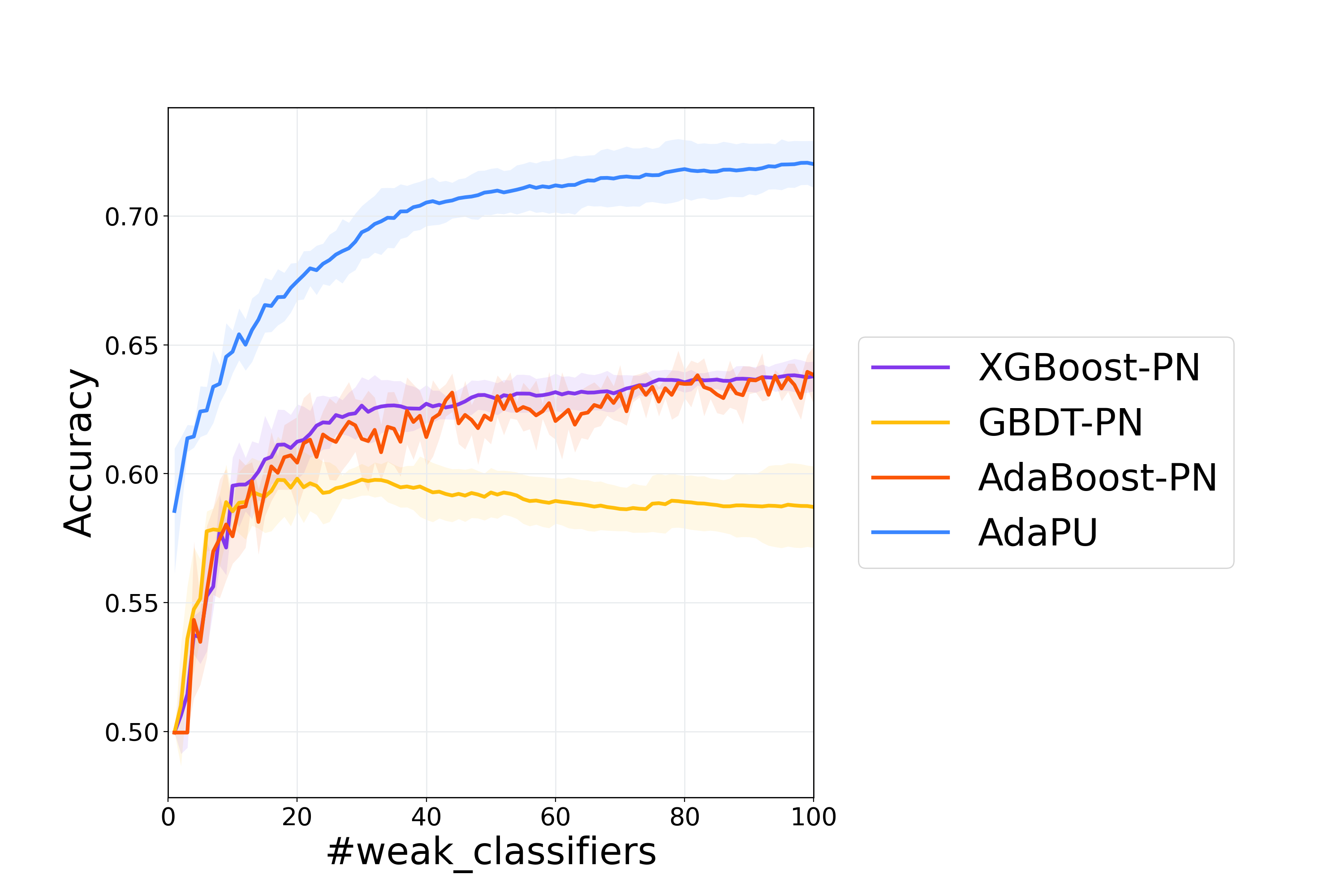}
        \end{minipage}%
    }%
    
    \subfigure[UNSW-NB15]{
        \begin{minipage}[htbp]{0.4\linewidth}
            \centering
            \includegraphics[width=.9\textwidth,height=.6\textwidth]{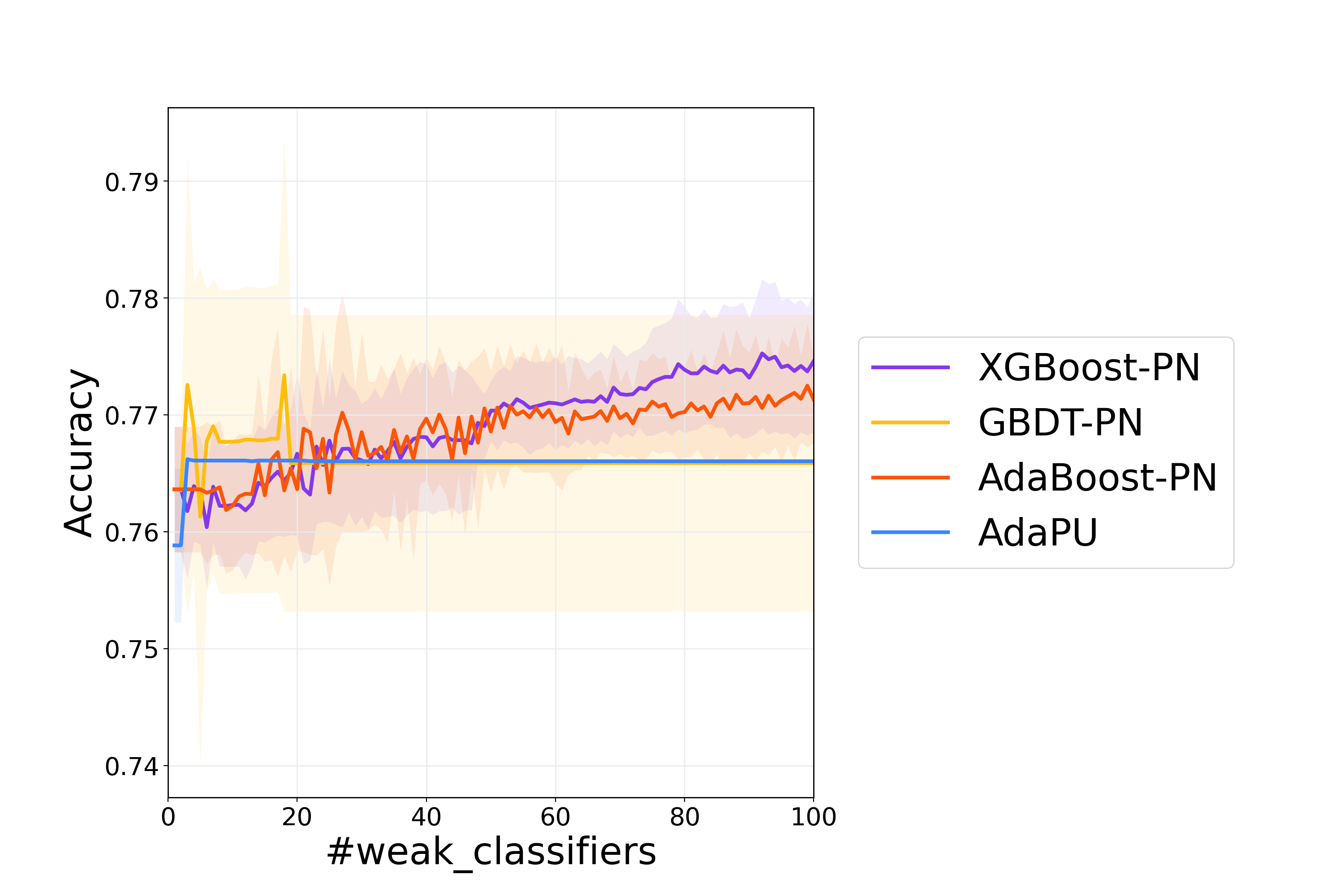}
        \end{minipage}%
    }%
    \subfigure[BreastCancer]{
        \begin{minipage}[htbp]{0.4\linewidth}
            \centering
            \includegraphics[width=.9\textwidth,height=.6\textwidth]{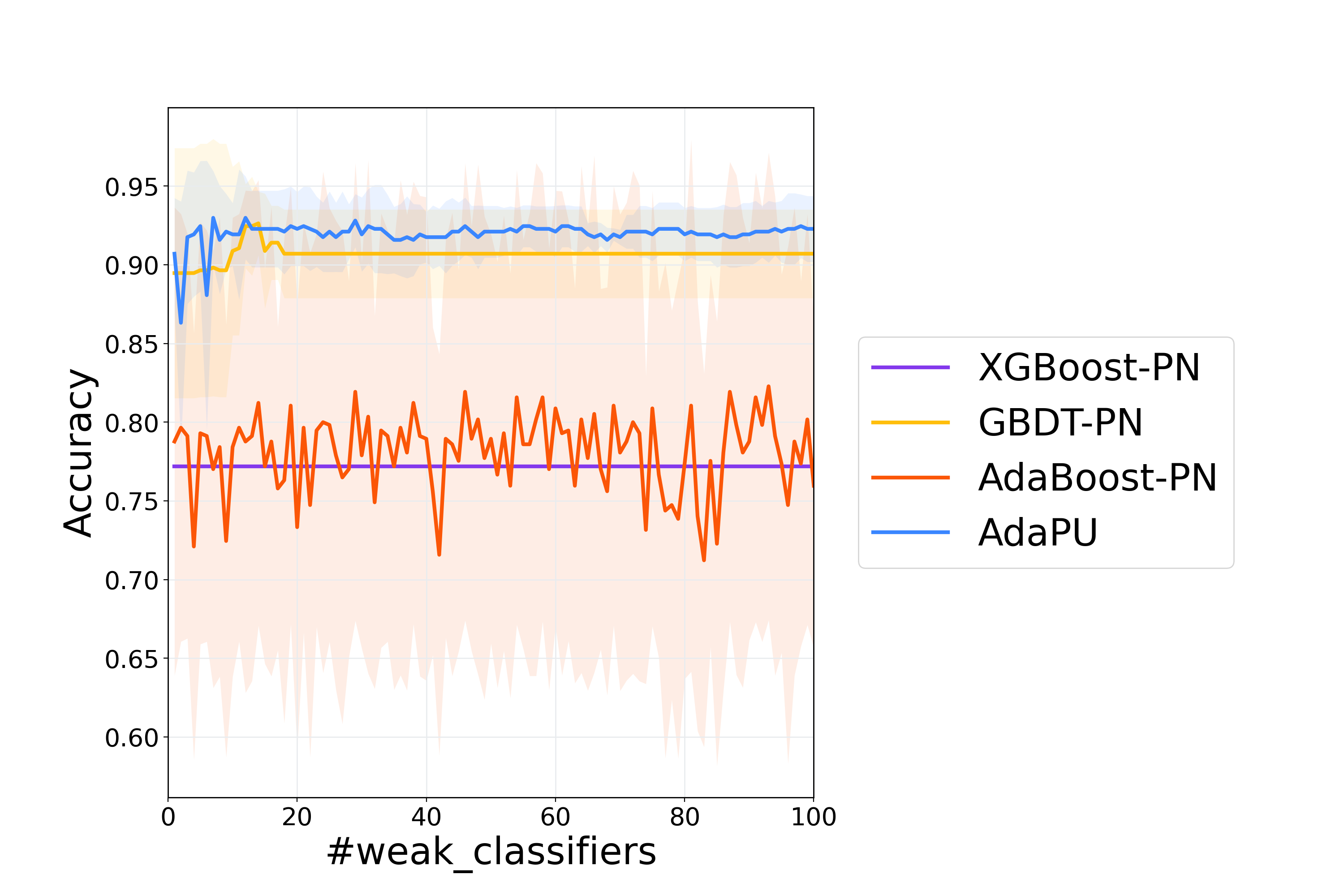}
        \end{minipage}%
    }%
    \centering
    \caption{Test accuracy of \codename~and PN boosting methods, on CIFAR-10,
		Epsilon, UNSW-NB15, and BreastCancer.
		The lines and shaded areas show the averages and standard deviations for 5 trials.
		}
    \label{fig:PNperformence}
\end{figure}

\medskip
\noindent \textbf{Comparison with PN boosting methods} 
The results of \codename~and PN boosting methods are shown in \Cref{fig:PNperformence} and \Cref{acc_table}. 
\codename performs better on Epsilon and BreastCancer, and is comparable to PN
boosting methods on CIFAR-10 and UNSW-NB15. 
This is consistent with the theoretical and experimental results in
\cite{niu2016theoretical}, which demonstrates that having a large amount of
unlabeled data may be better than having a small amount of negative data.

\begin{table}[t]
    \caption{\centering Accuracy in the form of the mean (std) for all compared methods. }
    \label{acc_table}
    \footnotesize
    \centering\setlength\tabcolsep{1mm}
    \begin{sc}
        \begin{tabular}{lccccc}
        \toprule
        & Method &  CIFAR-10 $\%$ &  Epsilon $\%$ & UNSW-NB15 $\%$ & BreastCancer $\%$ \\
        \midrule
        \multirow{2}{2cm}{PU Methods}& nnPU    & \textbf{87.17 (1.14)} & 70.38 (1.92) & 74.61 (0.26) & 90.35 (8.48)  \\
        & uPU         & 82.97 (4.42) & 61.53 (2.20) & 74.54 (0.49) & 76.84 (10.22)  \\
        \midrule
        
        \multirow{3}{2cm}{PN Methods}& XGBoost     & 85.29 (0.57) & 63.78 (0.58) & {\bfseries77.46 (0.60)} & 77.19 (0.00)   \\  
        & GBDT        & 86.46 (0.36) & 58.71 (1.58) & 76.59 (1.26) & 90.70 (2.82)  \\
        & AdaBoost    & 86.10 (0.54) & 63.83 (1.13) &  77.13 (0.36) & 75.96 (10.32) \\
        \midrule
        \multirow{2}{2cm}{AdaPU}& Over-All & 84.83 (0.13) & 71.60 (1.08) & 76.60 (0.0) & {\bfseries 93.86 (1.64)}  \\
        & Per-Group  & 85.77 (0.08) & {\bfseries 72.01 (0.91)} & 76.60 (0.0) & 92.28 (2.09)\\
        
        \bottomrule
        \end{tabular}
    \end{sc}
\end{table}

\subsection{Additional experiments on \codename}\label{ablation study}

\begin{figure}[ht]
    \centering
    \subfigure[CIFAR-10]{
        \begin{minipage}[t]{0.4\linewidth}
            \centering
            \includegraphics[width=.9\textwidth,height=.6\textwidth]{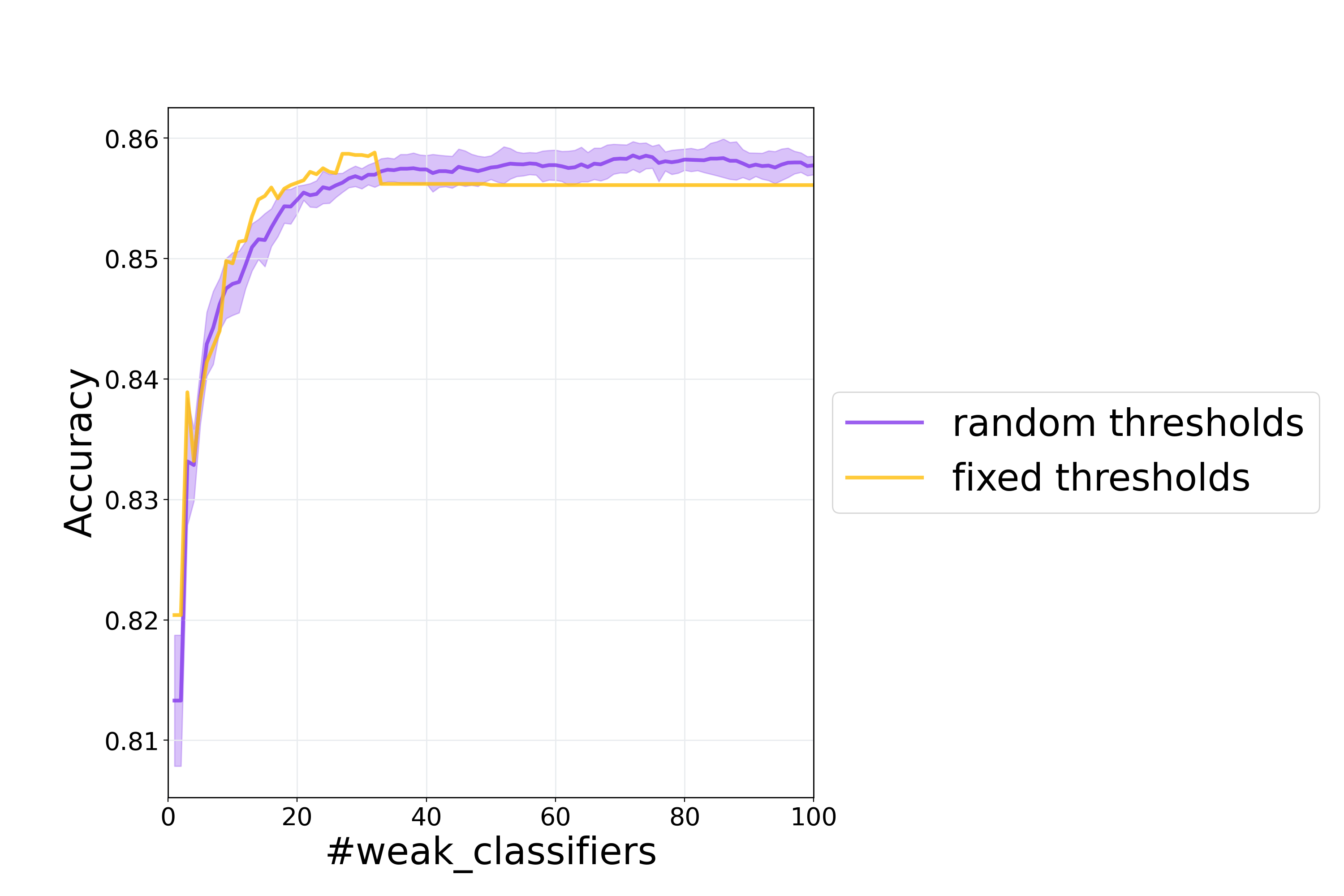}
        \end{minipage}%
        \begin{minipage}[t]{0.4\linewidth}
            \centering
            \includegraphics[width=.9\textwidth,height=.6\textwidth]{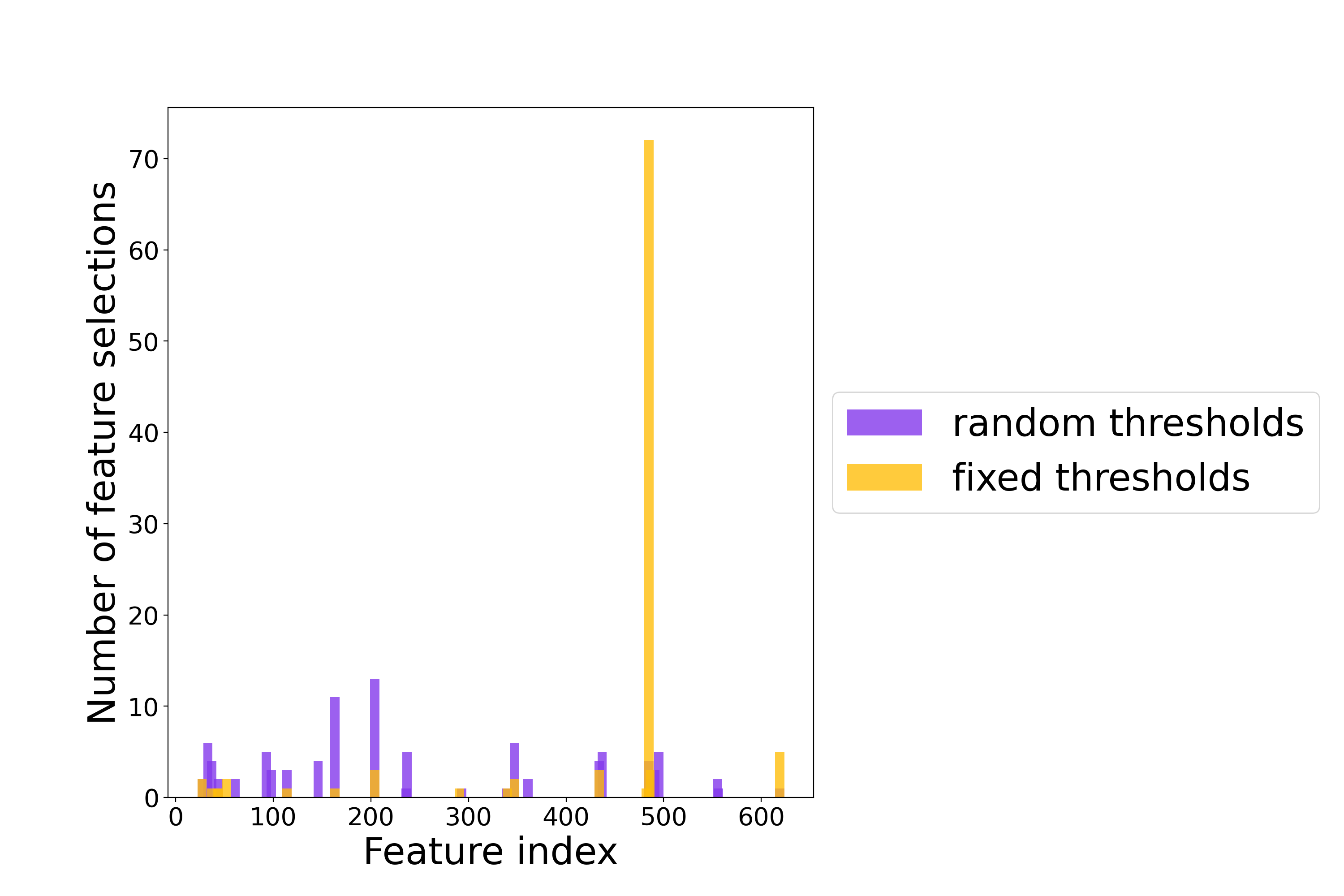}
        \end{minipage}%
    }%
    
    \subfigure[Epsilon]{
        \begin{minipage}[t]{0.4\linewidth}
            \centering
            \includegraphics[width=.9\textwidth,height=.6\textwidth]{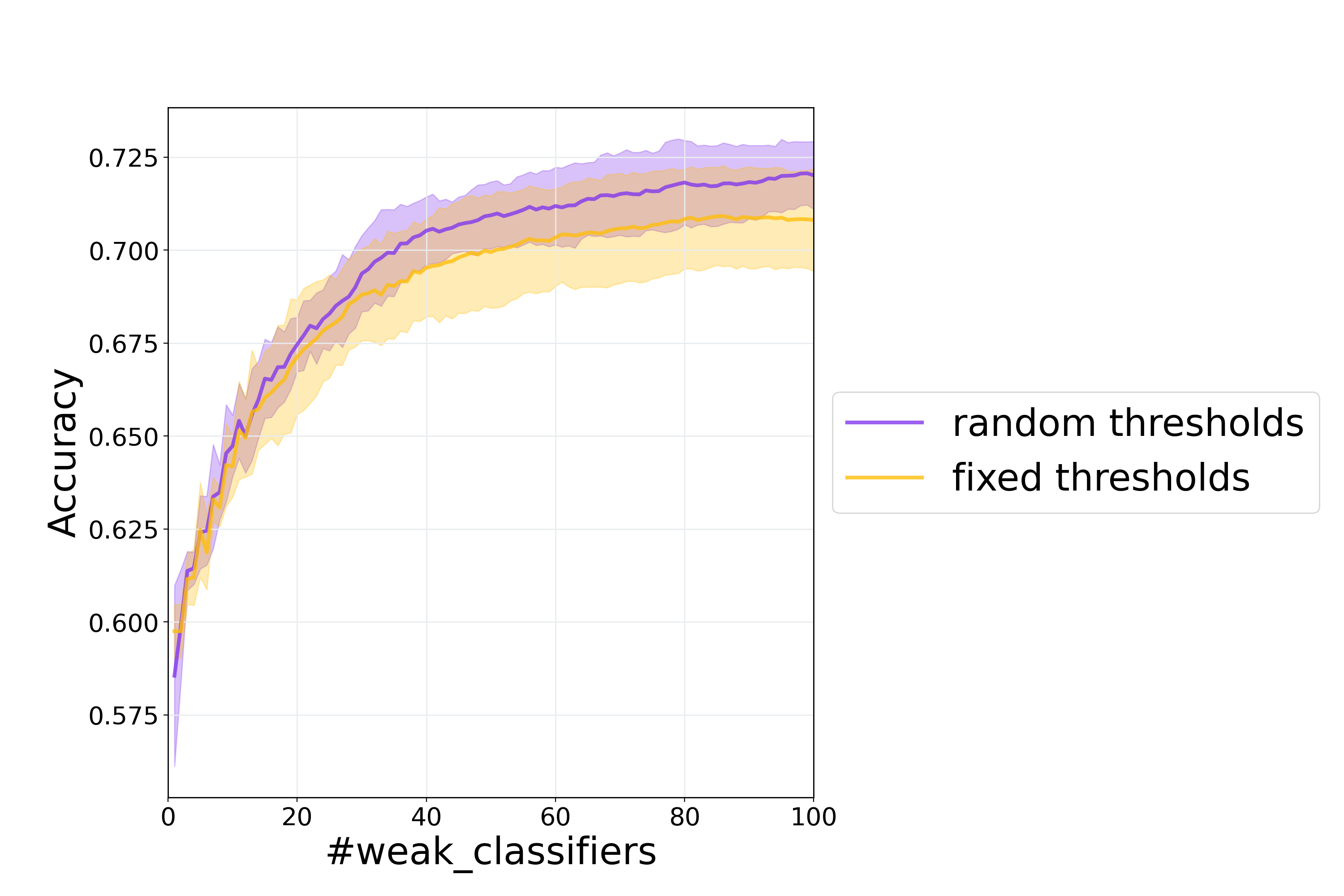}
        \end{minipage}%
        \begin{minipage}[t]{0.4\linewidth}
            \centering
            \includegraphics[width=.9\textwidth,height=.6\textwidth]{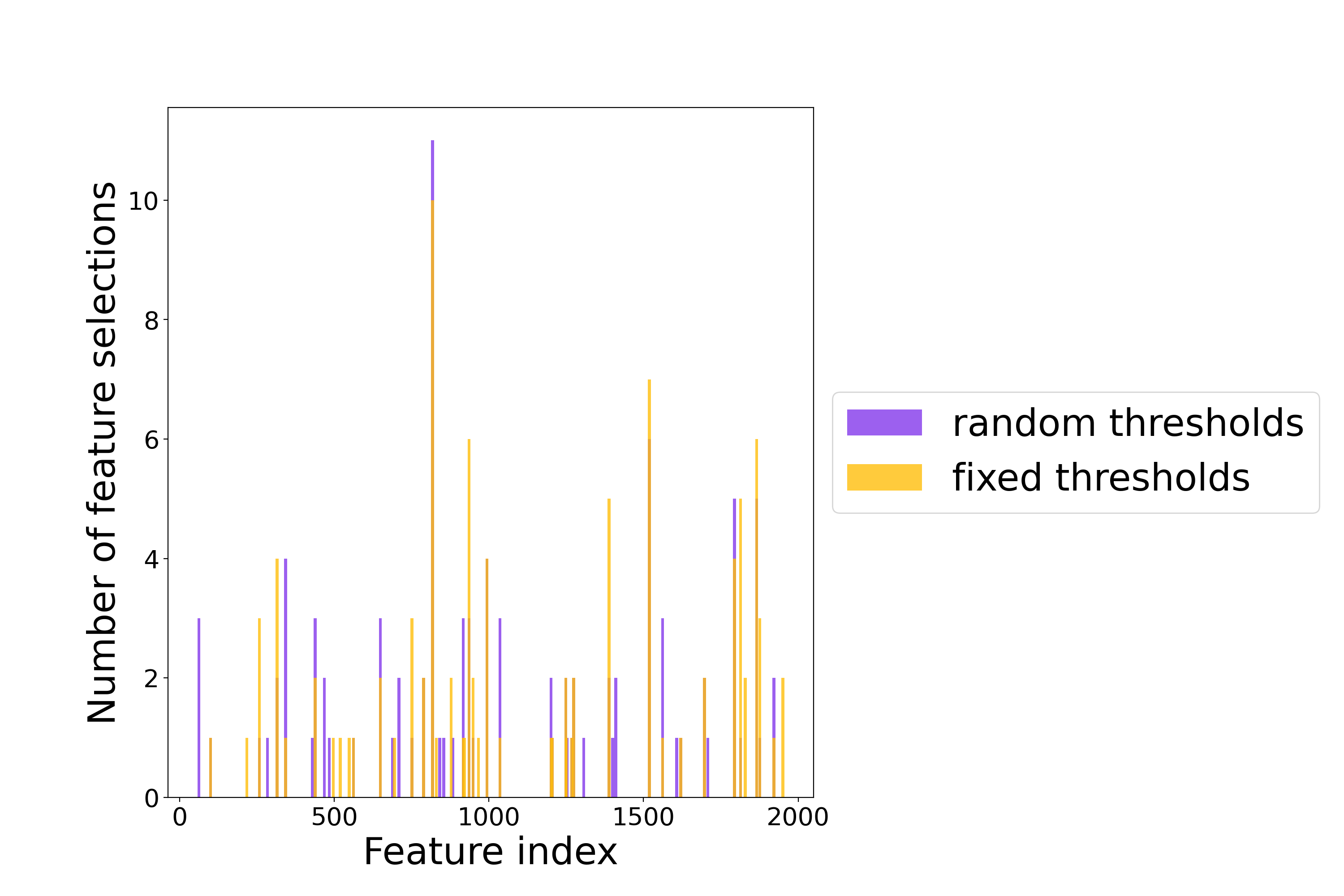}
        \end{minipage}%
    }%
		\caption{The test accuracies of \codename under different feature value selection strategies.
		The lines and shaded areas show the averages and standard deviations for 5 trials.
		The bar charts show the number of times features are used as splitting
		features in the trees in one trial.}
    \label{fig:random}
\end{figure}

{\bfseries Effects of feature threshold selection strategy.} 
\Cref{alg:stump} uses randomly sampled feature thresholds.
A natural alternative is to sample evenly spaced threshold values.
\Cref{fig:random} compares these two different threshold selection strategies on CIFAR-10 and Epsilon.
The random strategy leads to improved accuracies.
To understand how the different threshold selection strategies affect the performance 
of \codename, we plotted the number of times that a feature is used as a
splitting feature in the trees, for both datasets.
We can see that with randomly sampled thresholds, the number of times that features are selected
are more uniform, while with fixed thresholds, some features are selected much more frequently 
as compared to other features.
This suggests that spikes in feature frequencies may be related to overfitting and poorer 
generalization.

\begin{figure}[ht]
\centering

\subfigure[CIFAR-10]{
\begin{minipage}[htbp]{0.4\linewidth}
\centering
\includegraphics[width=.9\textwidth,height=.6\textwidth]{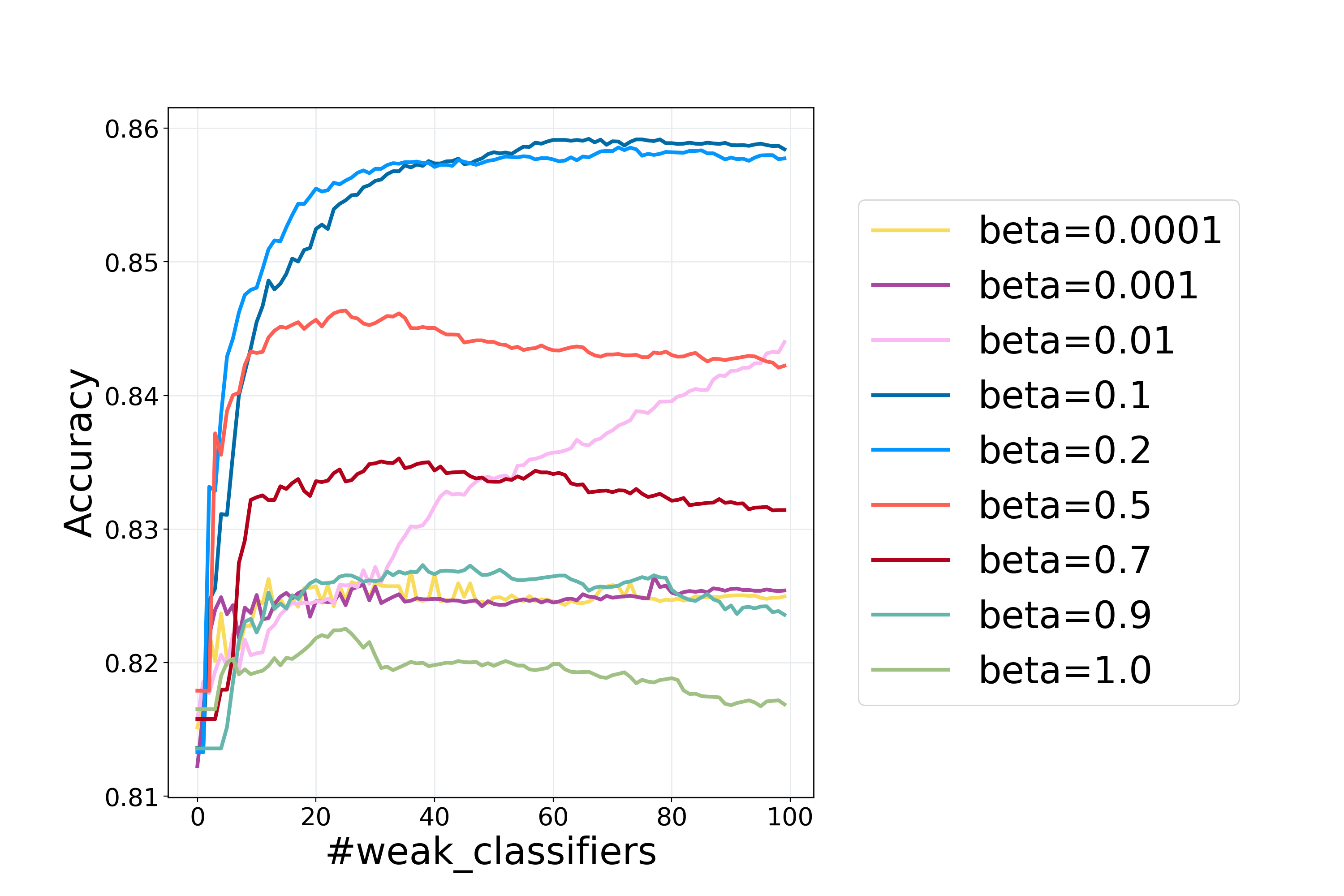}
\end{minipage}%
}%
\subfigure[Epsilon]{
\begin{minipage}[htbp]{0.4\linewidth}
\centering
\includegraphics[width=.9\textwidth,height=.6\textwidth]{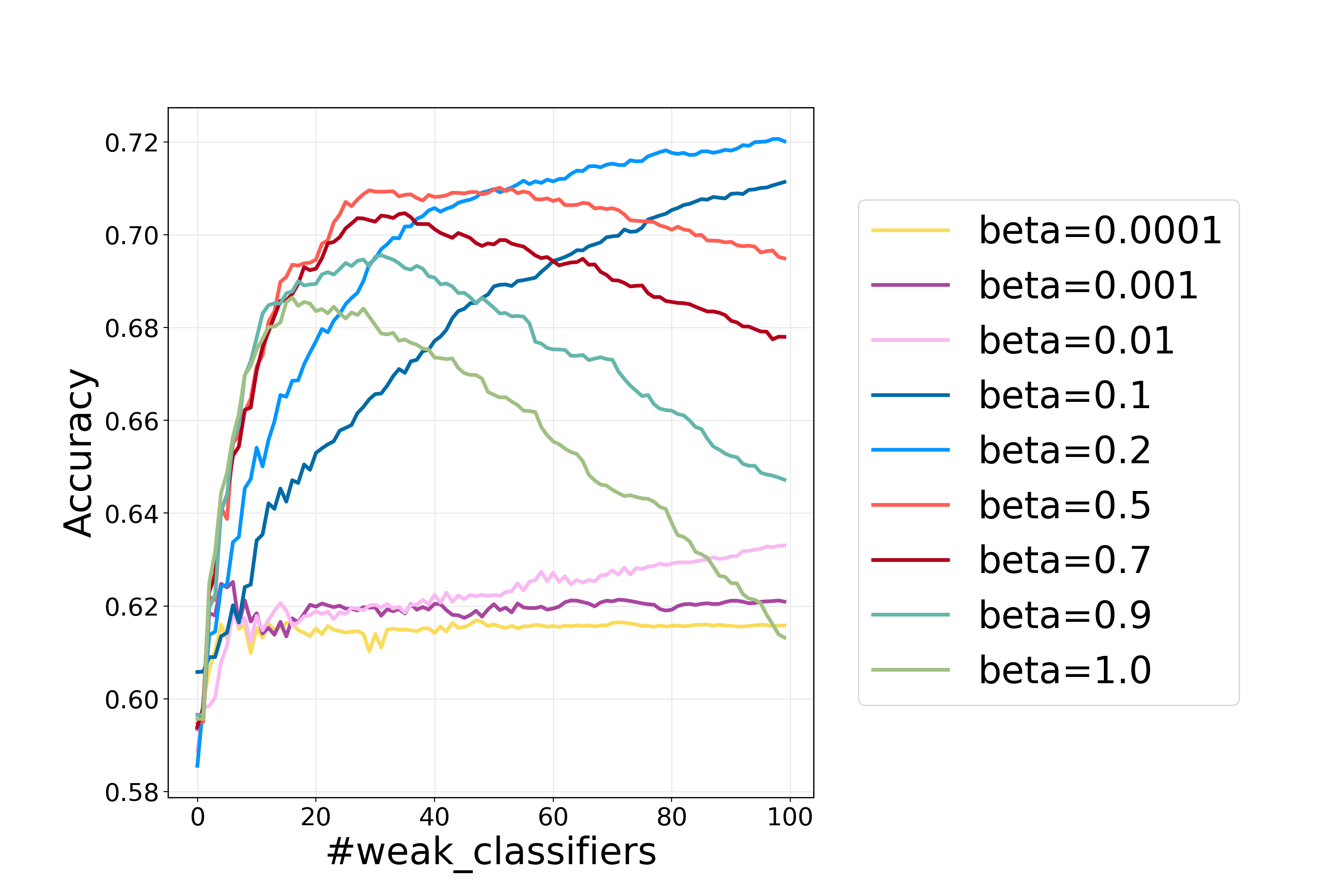}
\end{minipage}%
}%

\subfigure[UNSW-NB15]{
\begin{minipage}[htbp]{0.4\linewidth}
\centering
\includegraphics[width=.9\textwidth,height=.6\textwidth]{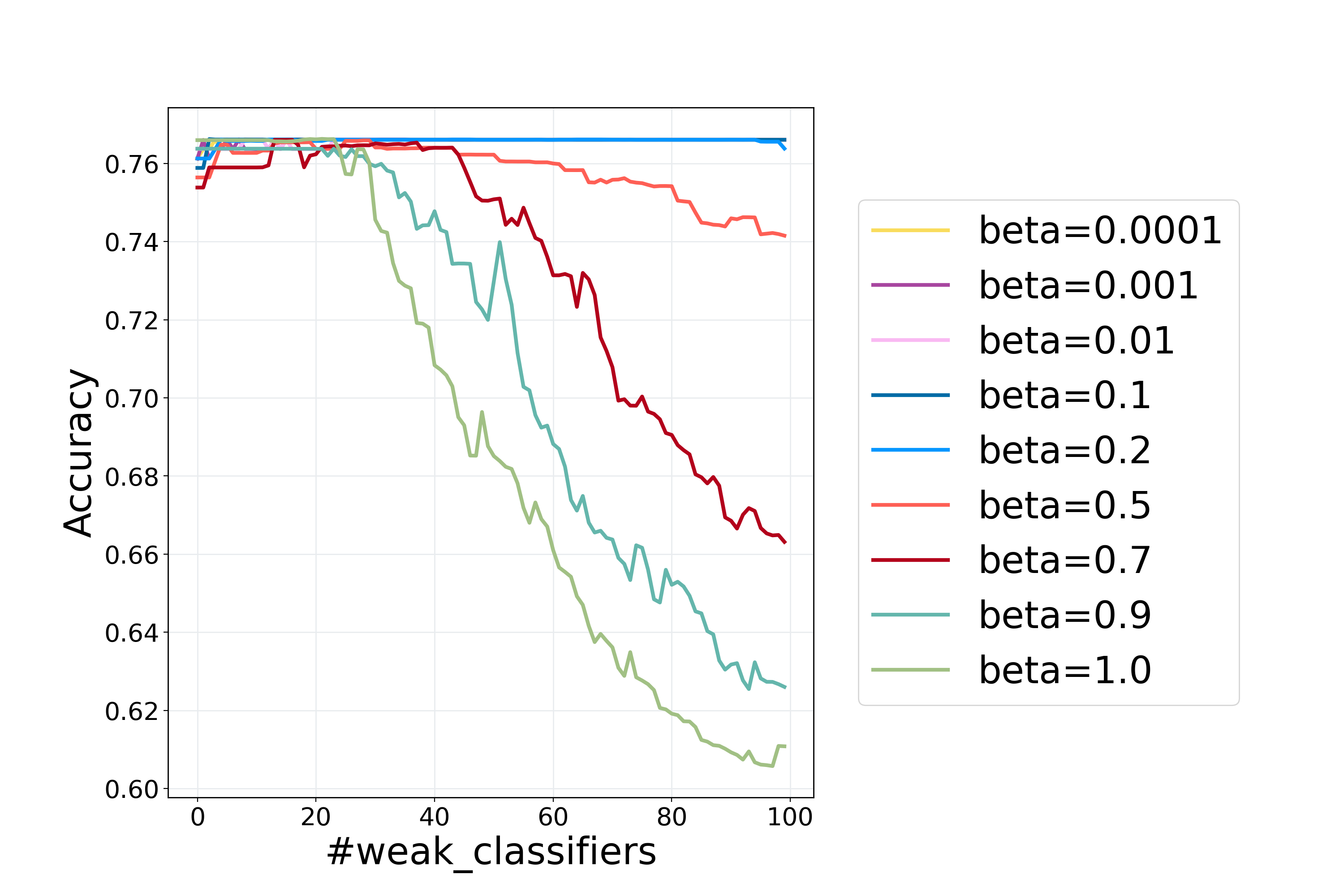}
\end{minipage}
}%
\subfigure[BreastCancer]{
\begin{minipage}[htbp]{0.4\linewidth}
\centering
\includegraphics[width=.9\textwidth,height=.6\textwidth]{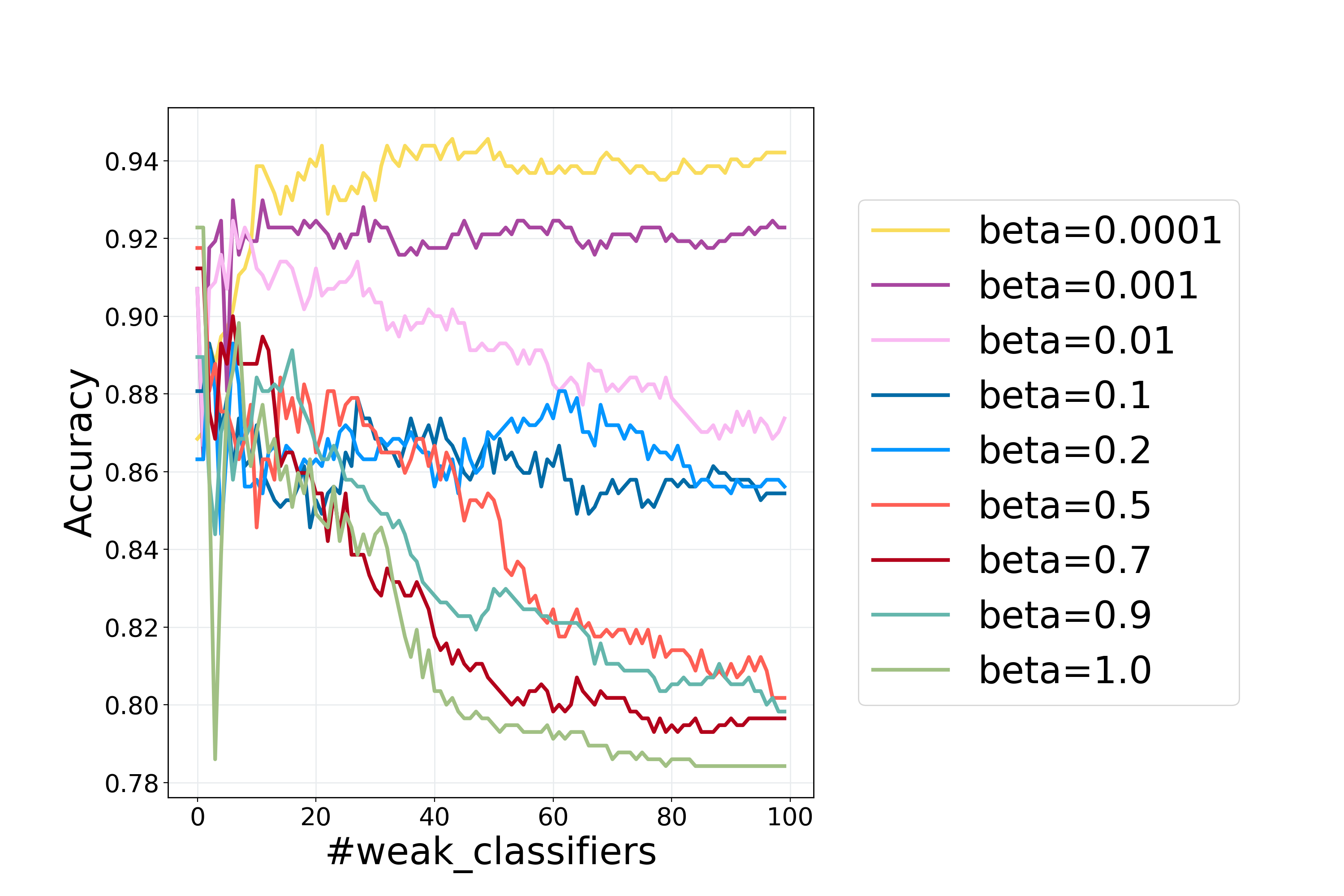}
\end{minipage}
}%

\centering
	\caption{Test accuracies of \codename for $\beta \in 0.0001, 0.001, 0.01, 0.1, 0.2, 0.5, 0.7, 0.9, 1.0\}$ 
	on CIFAR-10, Epsilon, UNSW-NB15 and BreastCancer. 
	The lines and shaded areas show the averages and standard deviations for 5 trials.
	}
\label{fig:lr}
\end{figure}

\medskip
\noindent {\bfseries Effects of $\beta$.} We explore how the $\beta$ affects the performance of \codename.
We evaluate the performance of \codename with $\beta \in \{0.0001, 0.001,0.01,0.1,0.2,0.5,0.7,0.9, 1.0\}$ in the experiment. 
\Cref{fig:lr} summarizes the comparison results of \codename~with different $\beta$ on CIFAR-10, Epsilon, UNSW-NB15 and BreastCancer. The results show that \codename has a better performance with a suitable $\beta$.

\begin{figure}[ht]
\centering

\subfigure[CIFAR-10]{
\begin{minipage}[htbp]{0.4\linewidth}
\centering
\includegraphics[width=.9\textwidth,height=.6\textwidth]{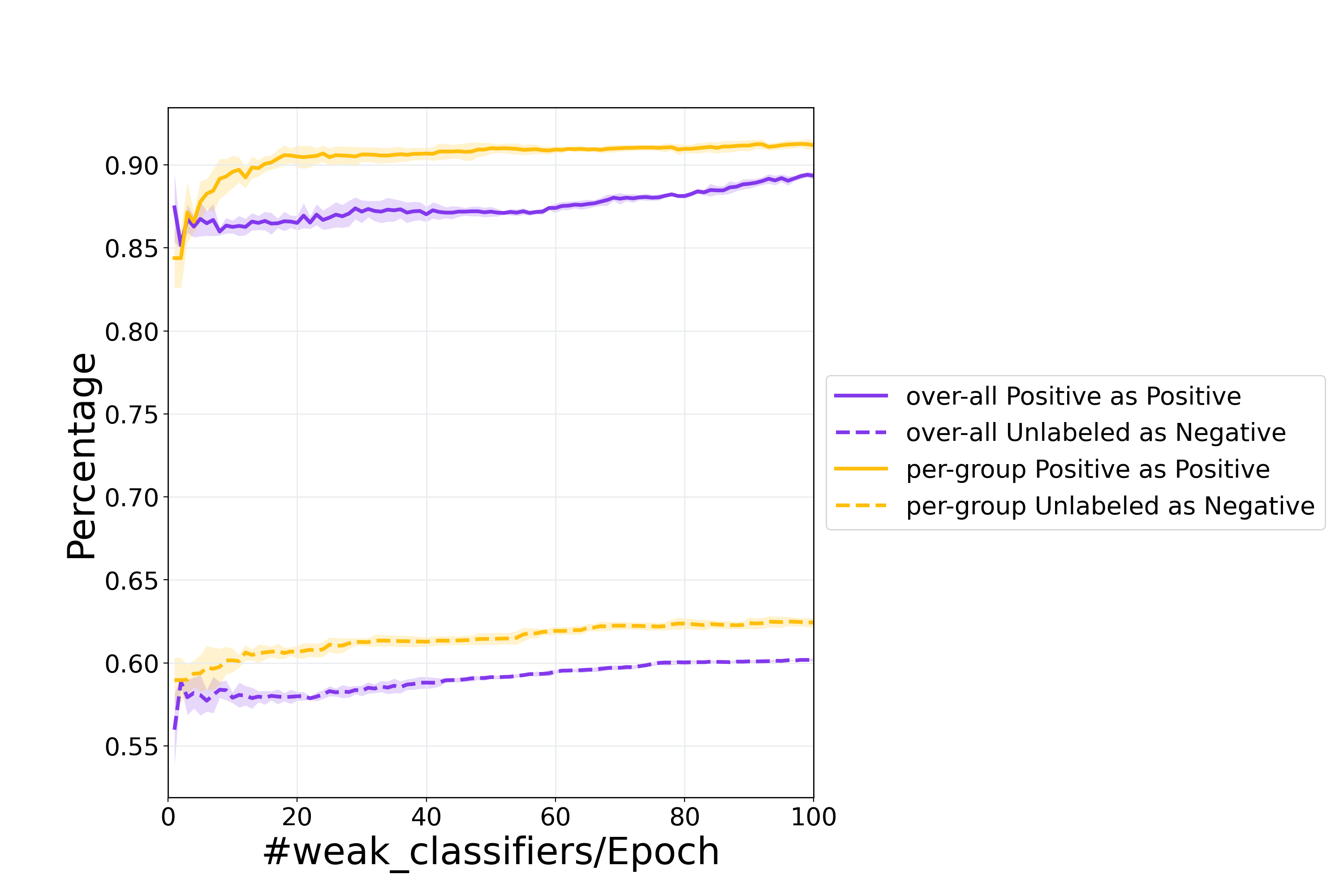}
\end{minipage}%
}%
\subfigure[Epsilon]{
\begin{minipage}[htbp]{0.4\linewidth}
\centering
\includegraphics[width=.9\textwidth,height=.6\textwidth]{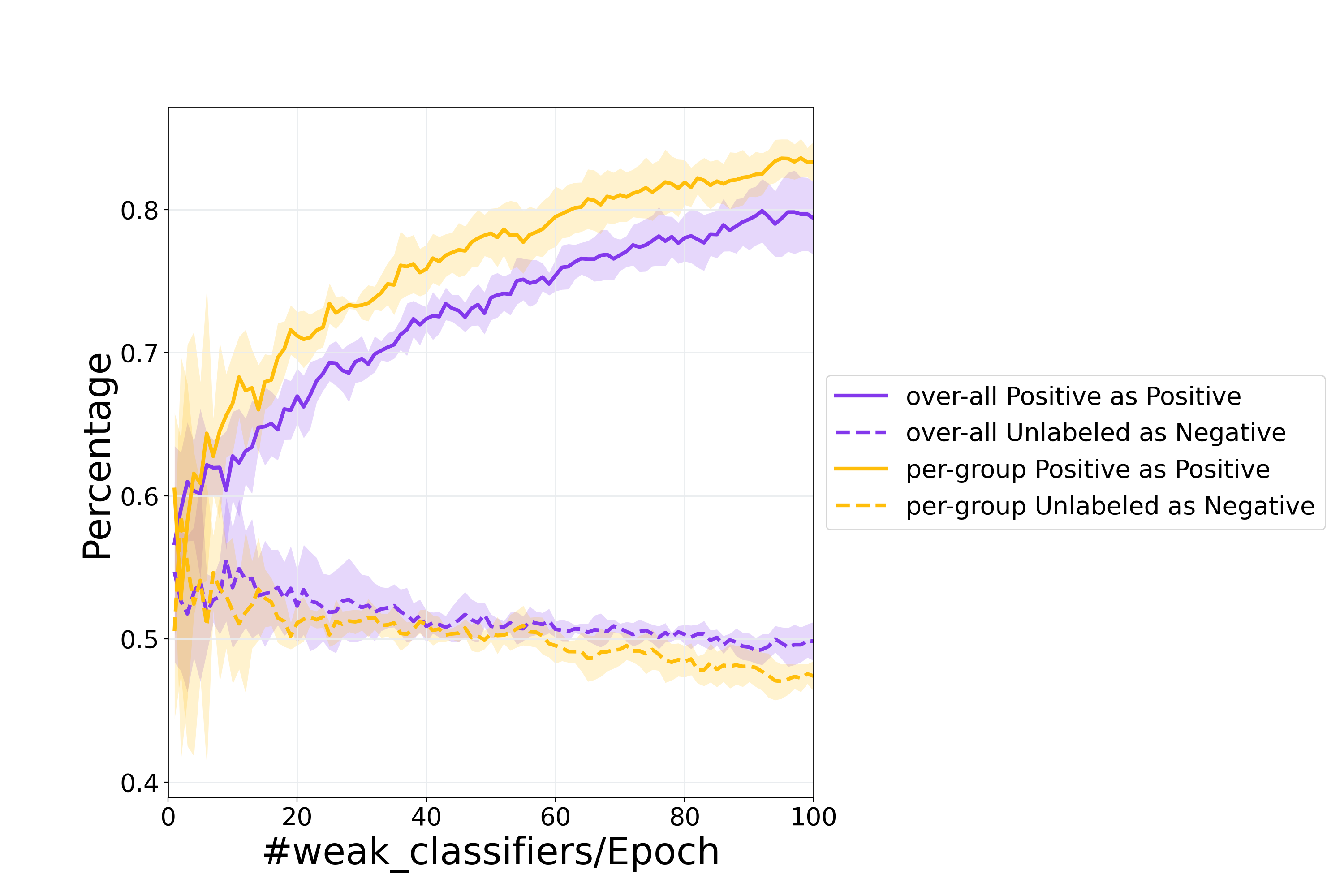}
\end{minipage}%
}%

\subfigure[UNSW-NB15]{
\begin{minipage}[htbp]{0.4\linewidth}
\centering
\includegraphics[width=.9\textwidth,height=.6\textwidth]{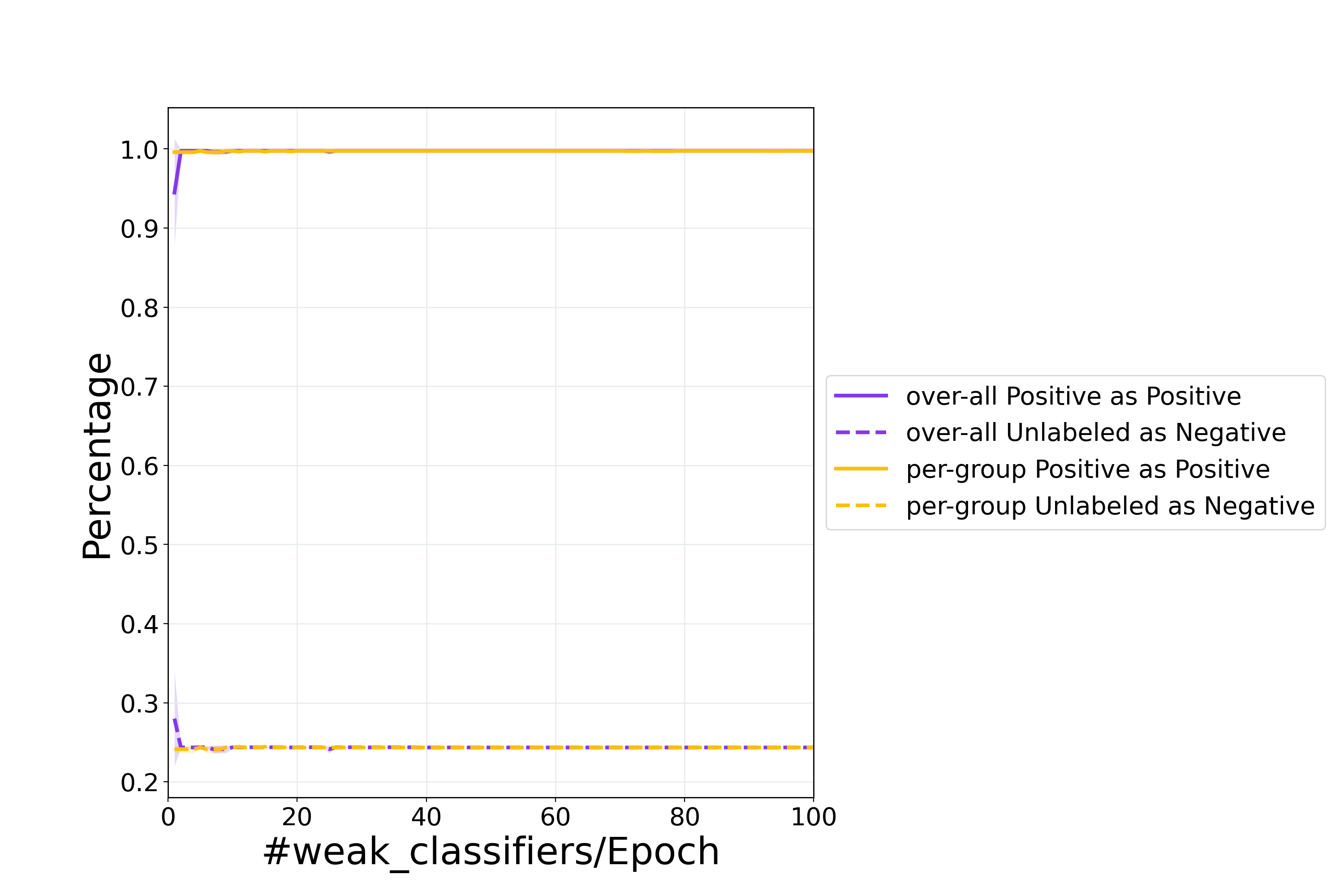}
\end{minipage}
}%
\subfigure[BreastCancer]{
\begin{minipage}[htbp]{0.4\linewidth}
\centering
\includegraphics[width=.9\textwidth,height=.6\textwidth]{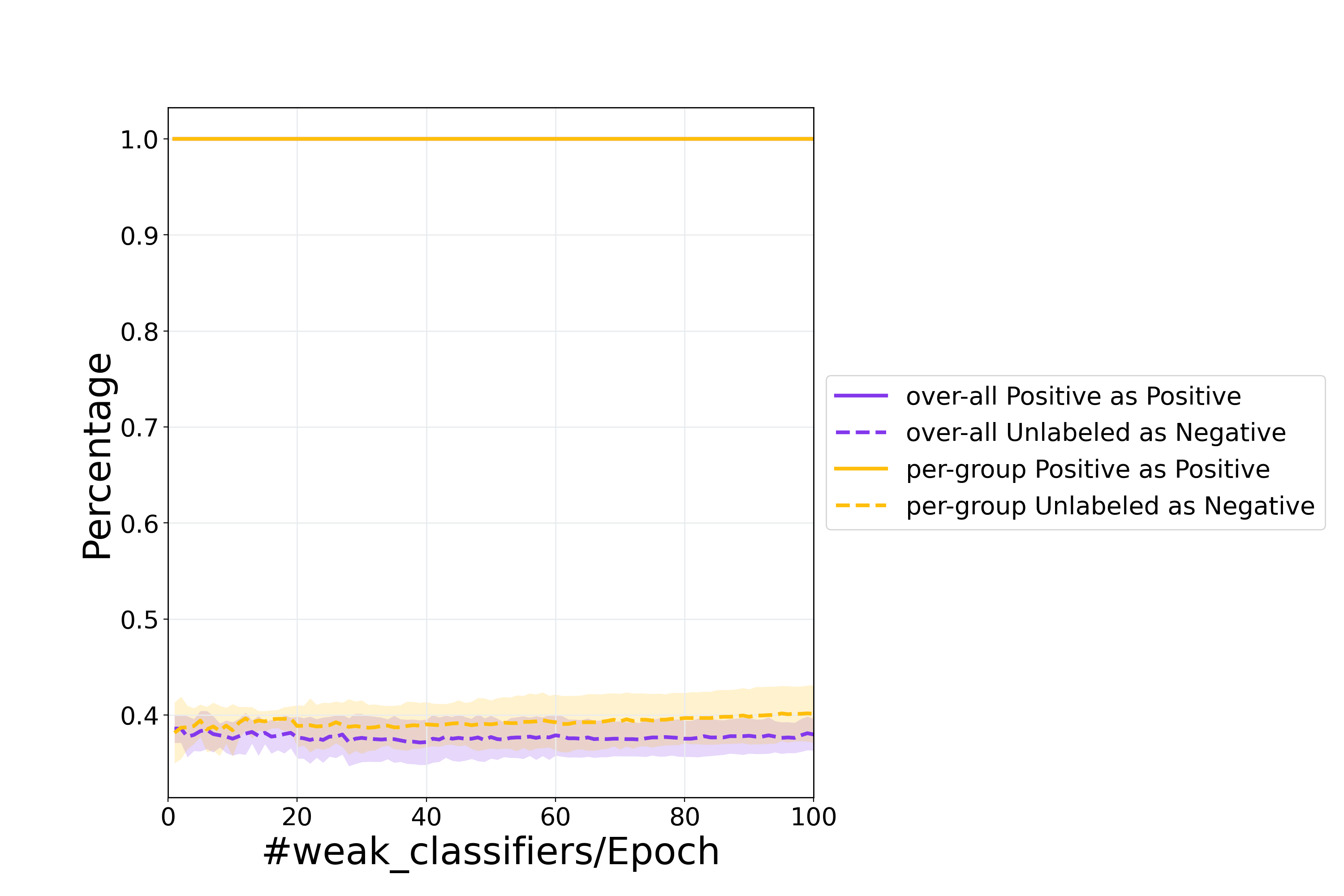}
\end{minipage}
}%

\centering
	\caption{Comparison of training set predictions on the positive and unlabeled examples for the 
	per-group normalization variant and the over-all normalization variant of \Cref{alg:stump}.
	The per-group normalization variant has a higher accuracy on positive examples in general.}
\label{fig:rate}
\end{figure}

\medskip
\noindent {\bfseries Effects of normalization method.}
As mentioned in \Cref{ada-PU section}, we experimented with two different ways of estimating a weak classifier $h$'s performance:
one with over-all normalization (i.e., using $e_{t}(h)$ to measure $h$'s classification error), the other
with per-group normalization (i.e., using $\epsilon_{t}(h)$ to measure $h$'s classification error).
We also used 5-fold cross-validation to select the $\beta$ values of \codename with over-all normalization. 
The selected $\beta$ values are as follows:
$\beta=0.1$ for Epsilon, $\beta=0.01$ for UNSW-NB15, $\beta = 0.0001$ for
BreastCancer, and $\beta = 0.01$ for CIFAR-10.
\begin{figure}[ht]
\centering

\subfigure[CIFAR-10]{
\begin{minipage}[htbp]{0.4\linewidth}
\centering
\includegraphics[width=.9\textwidth,height=.6\textwidth]{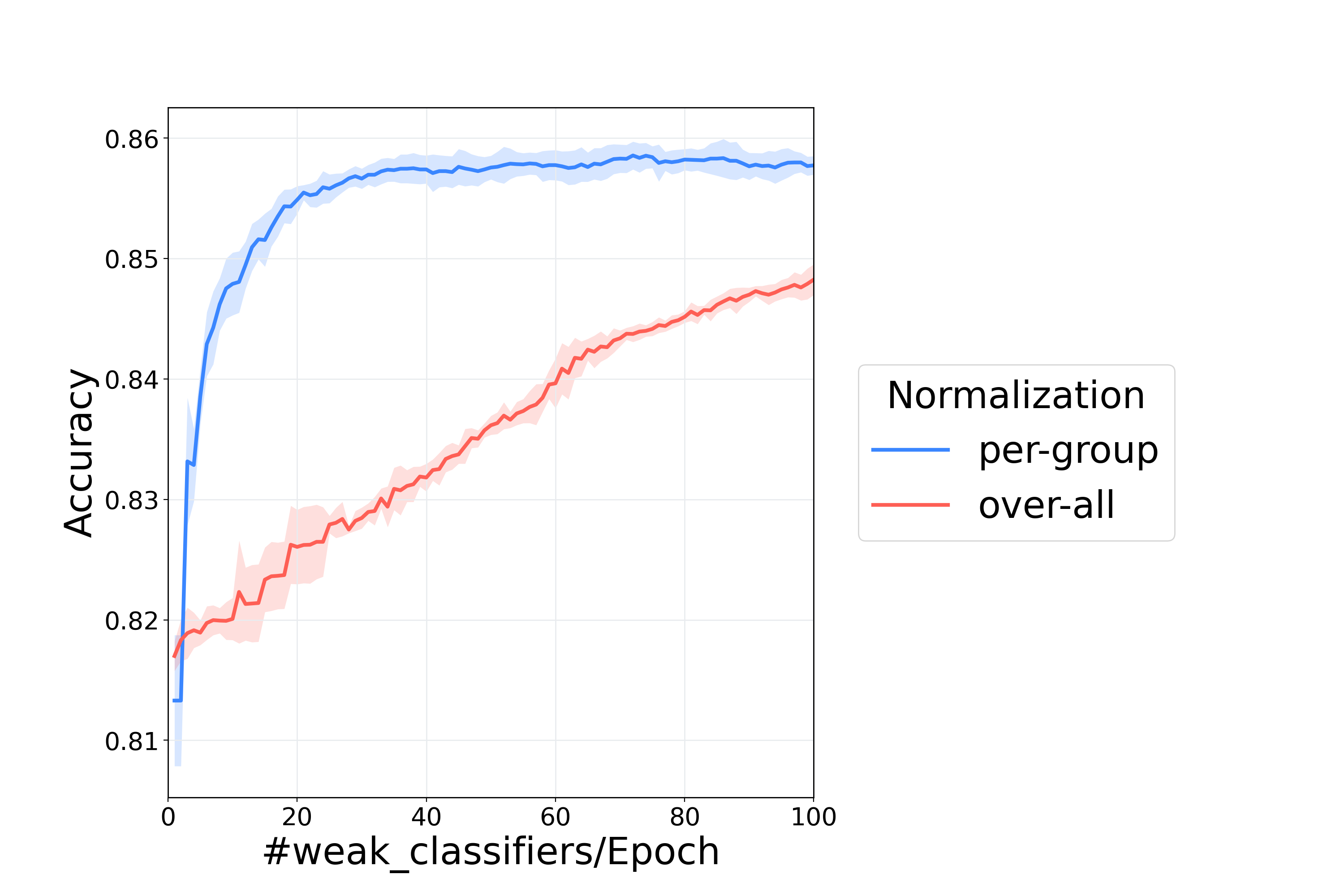}
\end{minipage}%
}%
\subfigure[Epsilon]{
\begin{minipage}[htbp]{0.4\linewidth}
\centering
\includegraphics[width=.9\textwidth,height=.6\textwidth]{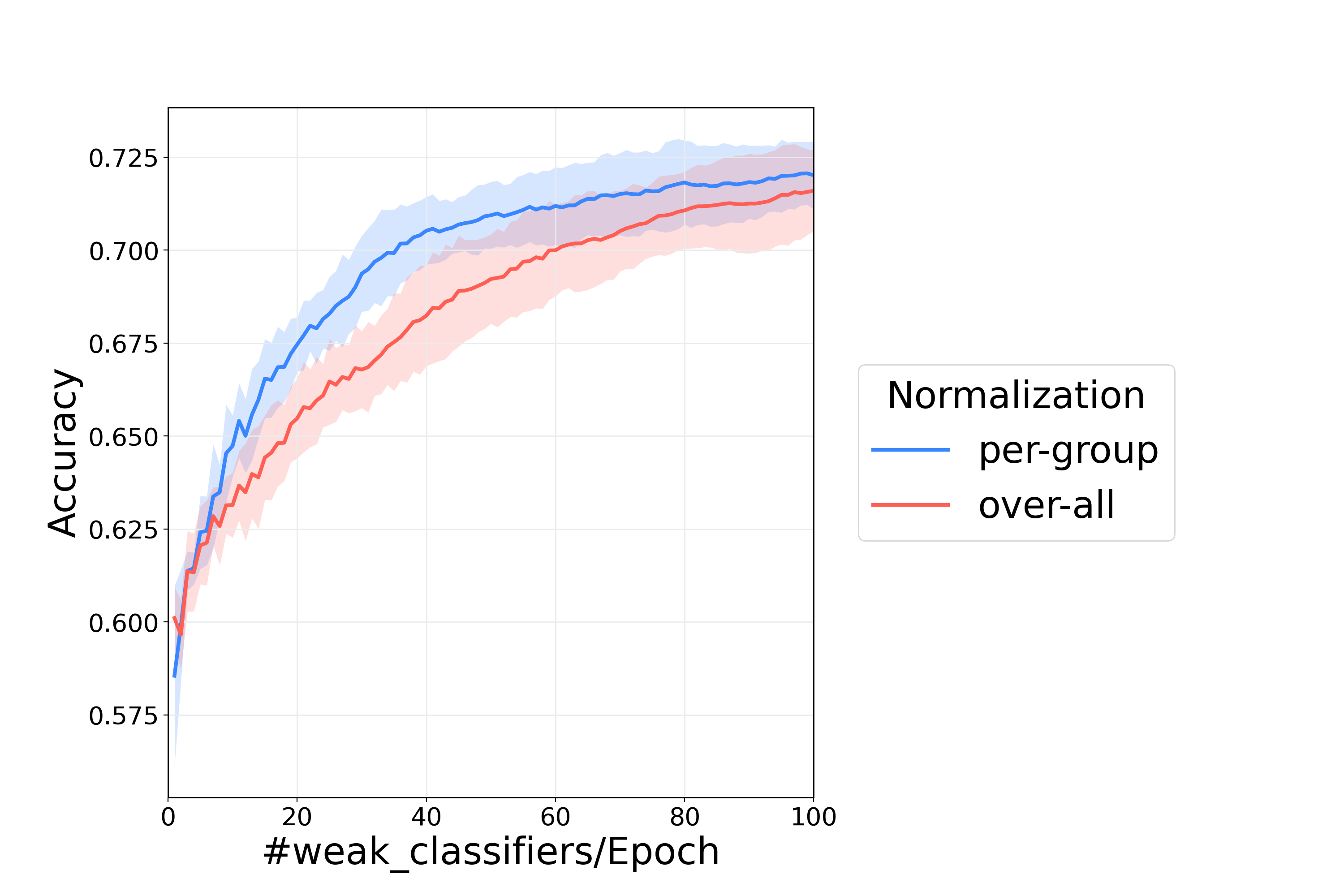}
\end{minipage}%
}%

\subfigure[UNSW-NB15]{
\begin{minipage}[htbp]{0.4\linewidth}
\centering
\includegraphics[width=.9\textwidth,height=.6\textwidth]{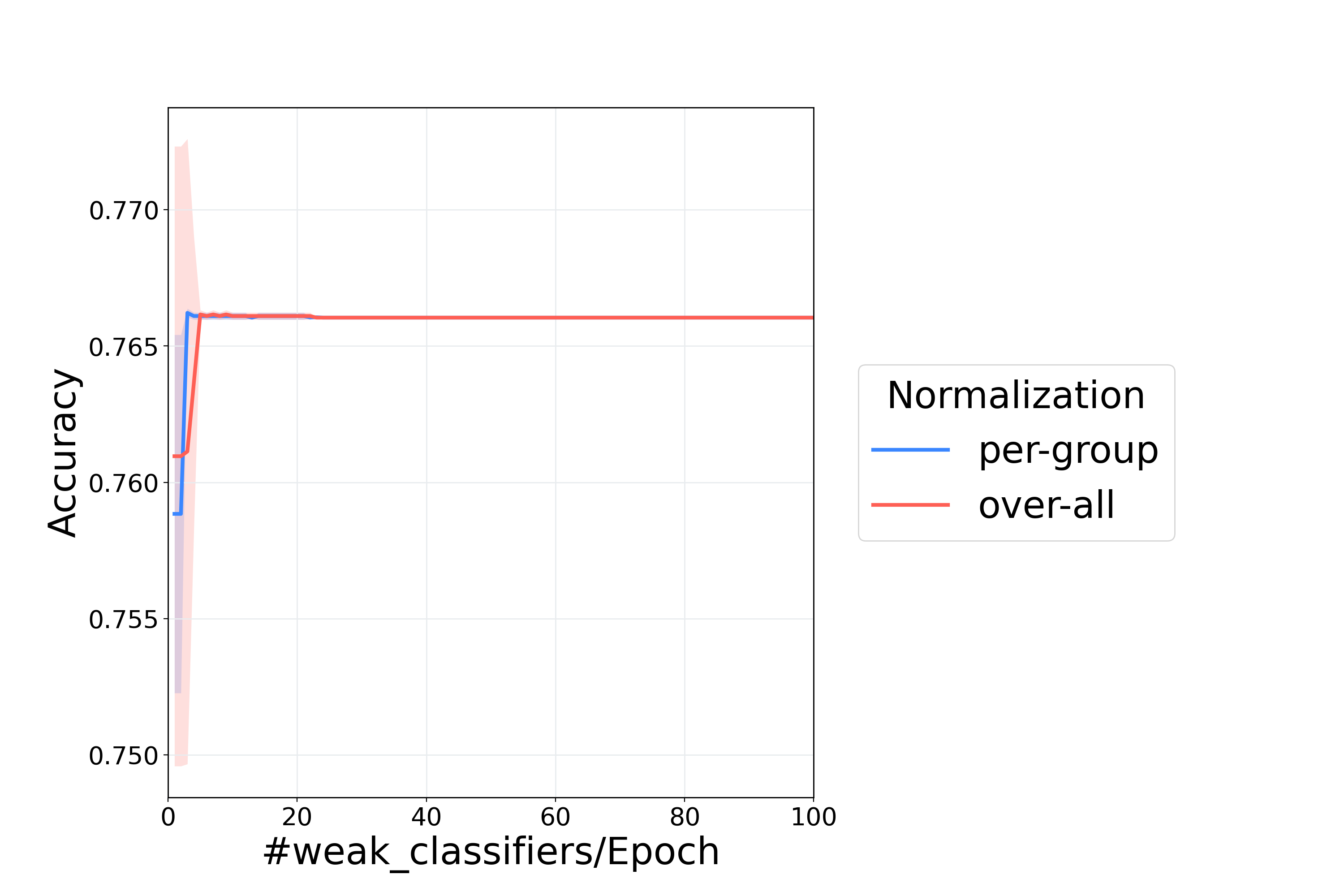}
\end{minipage}
}%
\subfigure[BreastCancer]{
\begin{minipage}[htbp]{0.4\linewidth}
\centering
\includegraphics[width=.9\textwidth,height=.6\textwidth]{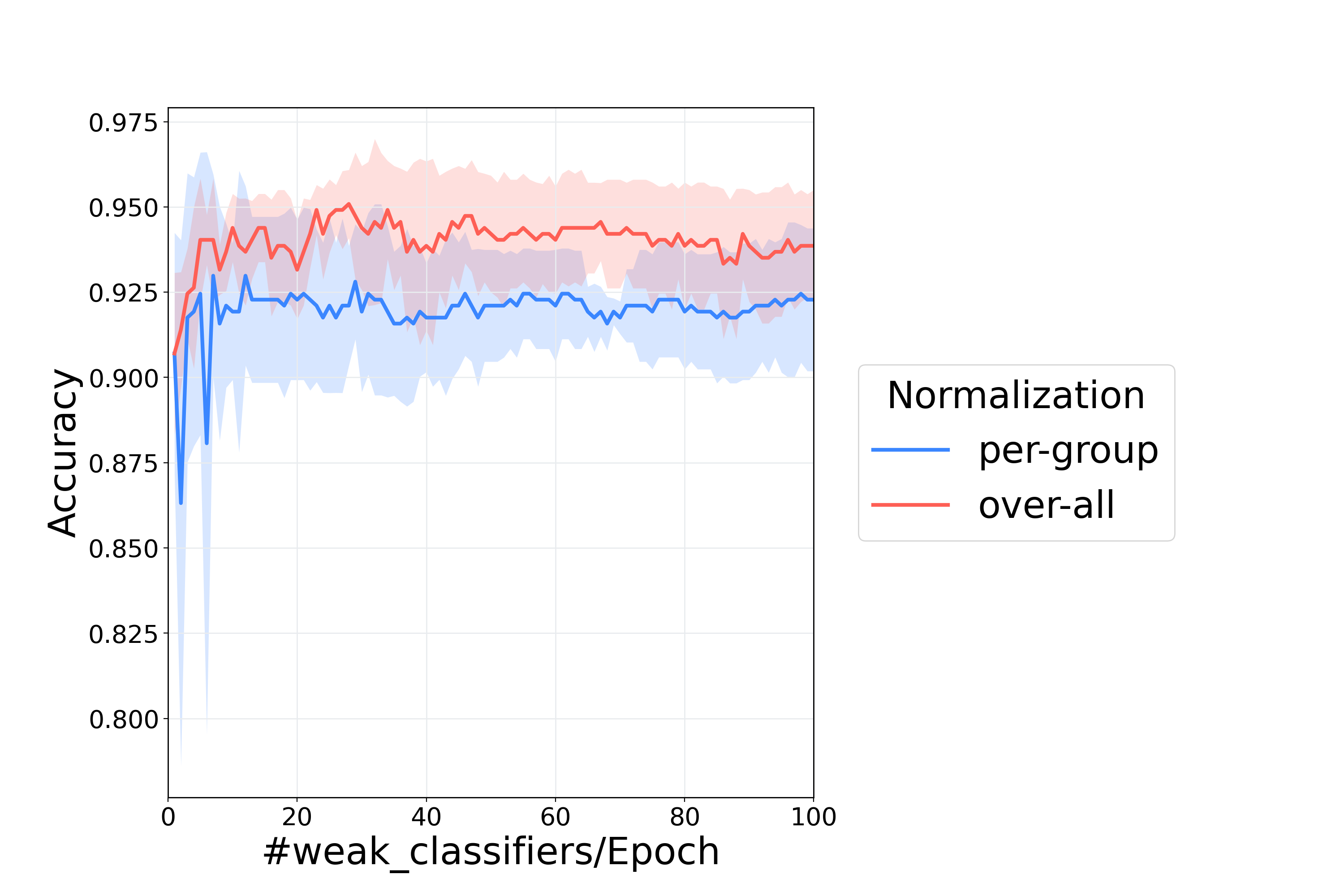}
\end{minipage}
}%

\centering
	\caption{Comparison of the test accuracies for the per-group normalization variant and the 
	over-all normalization variant of \Cref{alg:stump}.
	Per-group normalization is better than over-all normalization.}
\label{fig:norm}
\end{figure}
\Cref{fig:norm} shows that per-group normalization tends to have a faster convergence rate in general.
However, the final performance of both methods after training 100 weak classifiers are comparable, as shown in \Cref{acc_table}.
We further plotted the percentages of positives classified as positive and the percentages of unlabeled classified as negative,
as we add more classifiers, in \Cref{fig:rate}.
We can see that with per-group normalization, \codename tends to be more accurate in classifying positive examples,
particularly for CIFAR-10 and Epsilon.
At the same time, per-group normalization seems to have a mixed effect on \codename's tendency's to classify 
unlabeled examples as negative.
Overall, higher accuracy in classifying positive examples seem to be associated with faster learning, as 
seen on CIFAR-10 and Epsilon.
Given the same accuracy on positive examples, a method which is less aggressive in trying to classify 
unlabeled examples seem to perform better, as seen on BreastCancer.
This is likely a desirable behavior as discussed in \Cref{ada-PU section}, because unlabeled examples can be either positive or 
negative.

\section{Conclusion}\label{Conclusions}
We proposed a novel boosting PU learning method \codename in this paper. 
\codename shares some similarities with AdaBoost at a high level but
significantly differs from AdaBoost in how weak classifiers and their weights
are learned.
For computational efficiency, we focused on using decision stumps as the weak
classifiers.
The results show that \codename has strong performance on tabular data, even
with very simple weak classifiers.
It will be interesting to investigate the use of more complex weak classifiers
in \codename.

\FloatBarrier

\bibliographystyle{plainnat}
\bibliography{ref}

\end{document}